\pdfoutput=1

\documentclass[11pt]{article}

\usepackage{ACL2023}

\usepackage{times}
\usepackage{latexsym}

\usepackage[T1]{fontenc}

\usepackage[utf8]{inputenc}

\usepackage{microtype}

\usepackage{inconsolata}

\usepackage{multirow}
\usepackage{makecell}

\usepackage{times}
\usepackage{latexsym}
\usepackage{graphicx}
\usepackage{subfigure}
\usepackage{float}
\graphicspath{{images}}
\usepackage{stfloats}
\usepackage{booktabs}
\usepackage{indentfirst}
\usepackage{amsmath}
\usepackage{amssymb}

\usepackage{enumitem}

\usepackage{colortbl}
\usepackage{arydshln}

\usepackage[linesnumbered,ruled,vlined]{algorithm2e}

\SetCommentSty{mycommfont}

\SetKwInput{KwInput}{Input}                
\SetKwInput{KwOutput}{Output}              

\usepackage{CJKutf8}

\title{IAPT: \underline{I}nstruction-\underline{A}ware \underline{P}rompt \underline{T}uning for Large Language Models }

\author{
Wei Zhu$^1$$^\star$, \ 
Aaron Xuxiang Tian$^2$\thanks{\ \ Equal contributions. }, \ 
Congrui Yin$^3$, \ 
Yuan Ni$^4$$^\dagger$,  \\
\textbf{Xiaoling Wang$^1$\thanks{\ Corresponding authors: xlwang@cs.ecnu.edu.cn, niyuan442@pingan.com.cn.},} \ 
\textbf{Guotong Xie$^4$} \\ 
\textsuperscript{\rm 1} East China Normal University, Shanghai, China \\
\textsuperscript{\rm 2} Carnegie Mellon University \\
\textsuperscript{\rm 3} University of Minnesota \\
\textsuperscript{\rm 4} Pingan Technology, Shanghai, China  \\
}

\begin{document}
\maketitle
\begin{abstract}

Soft prompt tuning is a widely studied parameter-efficient fine-tuning method. However, it has a clear drawback: many soft tokens must be inserted into the input sequences to guarantee downstream performance. As a result, soft prompt tuning is less considered than Low-rank adaptation (LoRA) in the large language modeling (LLM) era. In this work, we propose a novel prompt tuning method, \underline{I}nstruction-\underline{A}ware \underline{P}rompt \underline{T}uning (IAPT), that requires only four soft tokens. First, we install a parameter-efficient soft prompt generator at each Transformer layer to generate idiosyncratic soft prompts for each input instruction. The generated soft prompts can be seen as a semantic summary of the input instructions and can effectively guide the output generation. Second, the soft prompt generators are modules with a bottleneck architecture consisting of a self-attention pooling operation, two linear projections, and an activation function. Pilot experiments show that prompt generators at different Transformer layers require different activation functions. Thus, we propose to learn the idiosyncratic activation functions for prompt generators automatically with the help of rational functions. We have conducted experiments on various tasks, and the experimental results demonstrate that (a) our IAPT method can outperform the recent baselines with comparable tunable parameters. (b) Our IAPT method is more efficient than LoRA under the single-backbone multi-tenant setting.

\end{abstract}

\begin{CJK*}{UTF8}{gbsn}

\section{Introduction}

Large language models (LLMs) have been emerging and achieving state-of-the-art (SOTA) results not only on a variety of natural language processing tasks \cite{qin2023chatgpt,PromptCBLUE,text2dt_shared_task,Text2dt,zhu_etal_2021_paht,Li2023UnifiedDR,Zhu2023BADGESU,Zhang2023LECOIE,Zhu2023OverviewOT,guo-etal-2021-global,zhu-etal-2021-discovering,Zheng2023CandidateSF,info:doi/10.2196/17653,Zhang2023NAGNERAU,Zhang2023FastNERSU,Wang2023MultitaskEL,Zhu2019TheDS,zheng2024nat4at,ACF,zuo-etal-2022-continually,Sun2020MedicalKG,li-etal-2019-pingan,zhu-etal-2019-panlp} but also many challenging evaluation tasks \cite{huang2023c,li2023cmmlu,Cui2023UltraFeedbackBL,yue2024tcmbench,yin2024machine,zhu2024text2mdt,text2dt_shared_task,zhou2019analysis,zhu2023overview} like question answering, reasoning, math, safety, instruction following. Despite LLMs becoming general task solvers, fine-tuning still plays a vital role in efficient LLM inference and controlling the style of the LLMs' generated contents.\footnote{Recently, OpenAI also released the fine-tuning API for GPT-3.5-turbo. See blog post: \url{https://openai.com/blog/gpt-3-5-turbo-fine-tuning-and-api-updates}.} Fine-tuning such large models by full parameters is prohibitive since it requires a large amount of GPU memory and computations. Thus, parameter-efficient fine-tuning (PEFT) \cite{Zhang2023LearnedAA,2023arXiv230318223Z,liu2024alora,zhu2023spt} has raised much attention in the research field since in PEFT, the tunable parameters are often less than 1\% of the LLMs and the computation costs will be significantly decreased.

\begin{figure*}
\centering
\includegraphics[width=0.78\textwidth]{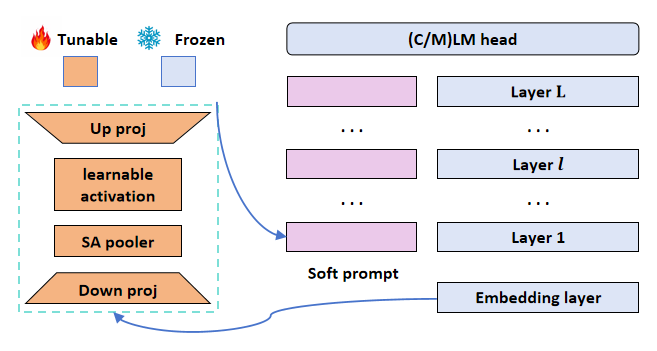}
\caption{Schematic illustration of our IAPT method. \textbf{Left}: The prompt generator which consists of a down-projection, a self-attention based pooler (SA pooler), a learnable activation whose curvature is learned in the downstream task, and a up-projection. \textbf{Right}: The prompt generator uses the instructions' hidden states as the input tensors, and output the generated soft tokens which will be concatenated to the next layer's hidden states.}
\label{fig:architecture}
\end{figure*}

Many PEFT methods have been validated to be effective across various models and tasks, often yielding comparable results with full-parameter fine-tuning \cite{He2021TowardsAU,zhu-tan-2023-spt,Zhang2023LearnedAA,zhu2023spt,liu2024alora,zhang-etal-2022-pcee,sun-etal-2022-simple}. Among these PEFT methods, the reparameterization-based method low-rank adaptation (LoRA) \cite{hu2021lora} is considered one of the most efficient and effective methods at present \cite{Xu2023ParameterEfficientFM,Ding2022DeltaTA,Xin2024ParameterEfficientFF,Zhu2021MVPBERTMP}. Although LoRA is effective and can bring stable performance with the original setting in \citet{hu2021lora}, it still brings inconvenience under the multi-tenant setting \cite{Chen2023PunicaML,zhu-etal-2021-gaml}: it has to add LoRA modules to multiple weights of the Transformer layer and introducing significant additional latency in every generation steps under the multi-tenant setting. Thus, it is of central importance to develop a novel PEFT method that introduces minimum latency during generation and still can perform competitively in downstream tasks.

In this work, we propose a novel PEFT method called \underline{I}nstruction-\underline{A}ware \underline{P}rompt \underline{T}uning (IAPT). We fine-tune the LLMs by inserting instruction aware soft prompts to each Transformer layer (Figure \ref{fig:architecture}). To flexibly regulate the attentions of LLMs, we install a prompt generator before each Transformer layer to generate the soft prompts by taking the input instruction's hidden states as input. The prompt generator is a lightweight module containing a down-projection layer, a self-attention based pooling layer, an activation function, and an up-projection layer. To enhance the expressiveness of prompt generators, we propose to automatically learn the idiosyncratic activation functions for different prompt generators with the help of rational functions.

We conduct extensive experiments on a wide collection of tasks, including sentiment classification, natural language inference, question answering, constrained sentence generation, math reasoning, SQL query generation, and instruction tuning, to demonstrate the effectiveness of our IAPT method. Notably, our method can consistently outperform strong PEFT baselines with comparable tunable parameter budgets, especially the recent LoRA variants \cite{Zhang2023AdaptiveBA,Ding2023SparseLA,liu2024alora} and SOTA prompt tuning methods \cite{Liu2022PTuningPT,Wu2022IDPGAI,Liu2022LatePT,zhu2023spt}. We also use experiments and analysis to show that: (a) our method has significantly lower latency under the multi-tenant setting than the LoRA-based methods with comparable tunable parameters. (b) We proposed adding a self-attention pooling module in the prompt generator, which can help the different transformer layers share the projection layers, thus improving the parameter efficiency. (c) The activation functions are learned during fine-tuning, which improves the downstream performance.

Our contributions are summarized as follows: 
\begin{itemize}
\item we propose a novel PEFT method, IAPT, that generates soft prompts conditioned on the input instructions received by the LLMs. 
\item We propose to employ the self-attention mechanism to obtain high-quality information aggregation of the input instructions, thus generating better soft prompts. 
\item We propose to learn the activation functions for different prompt generators at different Transformer layers, improving the downstream fine-tuning performance. 
\item We have conducted extensive experiments and analysis showing that our IAPT framework is (a) practical and outperforms the baselines under comparable tunable parameter budgets. (b) efficient during inference for LLMs. 
\end{itemize}

\section{Related works}

\subsection{Parameter-efficient fine-tuning (PEFT) methods}

Parameter-efficient fine-tuning (PEFT) is an approach of optimizing a small portion of parameters when fine-tuning a large pretrained backbone model and keeping the backbone model untouched for adaptation \cite{Ding2022DeltaTA,Zhang2023LearnedAA}.This branch of technology is of vital importance for applying (large) language models to downstream tasks with low resource consumptions \cite{Gao2023FPABEEFE,Guo2021GlobalAD,zhu-2021-leebert,zhang2021automatic,Wang2020MiningIH}. The addition-based methods insert additional neural modules or parameters into the backbone model. Representative works in this direction are Adapter \cite{houlsby2019parameter,Rckl2020AdapterDropOT,Zhang2023LearnedAA}, Prefix tuning \cite{li2021prefix}, Prompt tuning \cite{lester2021power}, P-tuning V2 \cite{Liu2022PTuningPT}. Another approach is called the specification-based approach, which is to specify the particular parameters to be tunable or prunable \cite{BenZaken2021BitFitSP,guo-etal-2021-parameter,zhao-etal-2020-masking}. The reparameterization-based methods have attracted much attention \cite{hu2021lora}. This branch of approaches transforms the adaptive parameters during optimization into low-rank and parameter-efficient forms. This type of PEFT method is motivated by the observation that fine-tuning has a low intrinsic dimension \cite{aghajanyan-etal-2021-intrinsic}. LoRA \cite{hu2021lora} hypothesizes that the change of weights during model tuning has a low intrinsic rank and optimizes the low-rank decomposition for the change of original weight matrices. PEFT methods are widely applied, especially with the popularization of open-sourced large language models \cite{2023arXiv230318223Z} and instruction tuning with these models for different application scenarios \cite{alpaca,2023arXiv230514314D}.

\subsection{Prompt tuning methods} 

Prompt tuning \cite{lester2021power} and P-tuning \cite{Liu2022PTuningPT} insert soft prompts to word embeddings only and can achieve competitive results when applied to supersized PTMs. Prefix-tuning \cite{li2021prefix} and P-tuning v2 \cite{Liu2021PTuningVP} insert prompts to every hidden layer of PTMs. IDPG \cite{Wu2022IDPGAI} uses parameterized hypercomplex multiplication \cite{Le2021ParameterizedHG} to parameterize soft prompts, improving the parameter efficiency. LPT \cite{Liu2022LatePT} improves upon IDPG by selecting an intermediate layer to insert soft prompts. SPT \cite{zhu-tan-2023-spt} designs a mechanism to automatically decide which layers to insert new soft prompts or keep the prompts propagated from the previous layer. Our work is different and compliments the existing literature in the following aspects: (a) The above works do not work with the current SOTA large language models and only experiment with relatively simple classification tasks. In comparison, by generating soft prompts conditioned on the input instructions received by the LLMs, our IAPT method works well with the currently best decoder-based LLMs in a wide collection of downstream tasks. (b) Our work can reduce the number of soft tokens from 32-128 to 4 by improving the architectural design of the prompt generators. (c) our work improves the parameter efficiency by sharing the parameters of prompt generators across Transformer layers.

\section{Methods}

\subsection{Preliminaries}

\noindent \textbf{Transformer model} \quad Currently, the most widely used open-sourced (large) language models adopt the stacked Transformer architecture \cite{Vaswani2017AttentionIA}. Denote the total number of Transformer layers in the LLM as $L$. Each Transformer block is primarily constructed using two key submodules: a multi-head self-attention (MHA) layer and a fully connected feed-forward (FFN) layer. 

\noindent \textbf{Task format} \quad Denote the task's training set as $\mathcal{D}_{\text{train}} = {(x_m, y_m), m = 1, 2, ..., M}$, where $M$ represents the number of samples. In this work, we only consider the case where input $x_m$ and target $y_m$ are both text sequences. Moreover, we expect the language modeling head of LLMs to decode $y_m$ during inference. That is, no additional linear prediction heads are considered for predicting categorical or numerical values.

\subsection{Motivation}


Soft prompt tuning is a widely studied PEFT method, which can achieve downstream task adaptations through a minimal number of tunable parameters compared to the (L)LM backbones. However, soft prompt tuning is less applied or studied in the era of large language models \cite{Xu2023ParameterEfficientFM,Ding2022DeltaTA,Xin2024ParameterEfficientFF} due to the following drawbacks: 
\begin{itemize}
\item Soft prompt tuning needs to concat a sequence of soft prompts to the input sequence, which inevitably increases the model complexity significantly. According to the experiments in \cite{PromptCBLUE}, P-tuning V2 \cite{Liu2022PTuningPT} has to set the soft prompts' sequence length to at least 32 to make the LLM's downstream task performance to be close to LoRA \cite{hu2021lora}.   
\item Soft prompt tuning can not flexibly adjust the tunable parameter numbers if the soft prompt length is fixed, thus can not conveniently make a tradeoff between the PEFT parameters' expressiveness and the parameter efficiency.

\item Most existing work on soft prompt tuning methods assumes that the soft prompt is fixed and shared across all samples within a task or dataset. However, a task may encompass a diverse range of samples, some of which are easy while others are challenging. Therefore, prompts need to exhibit sufficient diversity to effectively modulate the attention patterns of LLMs across different samples, thereby achieving better fine-tuning performance.

\end{itemize}

Thus, in this work, we propose dynamically generating prompt generators based on the given instructions/prompts to address the above issues. 


\subsection{Instruction aware prompt generators}

In this work, the prompt generators generate soft prompts based on the input instructions received by the LLMs. As shown in Figure \ref{fig:architecture}, in order to generate responses, the input instructions have to go through the LLM backbone to obtain the hidden representations. Denote the hidden state of the input instruction with length $l_{ins}$ right before the current Transformer layer as $\mathbf{h}$.\footnote{Note that the soft prompts propagated from the previous layer will be discarded. That is, each Transformer layer uses newly generated prompts. We will use experiments to demonstrate the validity of this setting. } The prompt generator first down-projects $\mathbf{h}$ from dimension $d$ to dimension $m$ ($m \ll d$) via a linear layer $\text{MLP}_{down}$. Then, it obtains the prompt $\mathbf{p}$ with a fixed length $l_{sp}$ through a pooling operation $\text{Pooler}()$. The pooled prompt will go through an activation function $g$ and be up-projected to dimension $d$ via another linear layer $\text{MLP}_{up}$. Formally, 
\begin{equation}
\mathbf{p} = \text{MLP}_{up}(g(\text{Pooler}(\text{MLP}_{down}(\mathbf{h})))).
\label{eq:prompt_generator}
\end{equation}
Then, the generated soft prompt $\mathbf{p}$ will be concatenated to $\mathbf{h}$ and go through the calculations of the next Transformer layer.

Note that the decoder-based causal language models (CLM) usually employ the KV cache mechanism\footnote{See the blog post for an in-depth explanation of KV-cache: \url{https://www.dipkumar.dev/becoming-the-unbeatable/posts/gpt-kvcache/}. } during generation to reduce computation complexity. Our prompt generators work seamlessly with the KV cache mechanism since the soft prompts are generated when the input instruction (or prompt) is passed through the LLM for the first time. In the subsequent generation steps, the generated soft prompts will be combined into the KV caches and will be reused without repetitively calling the prompt generators. In comparison, the LoRA method provides reparameterizations to the model parameters, and it has to participate in the calculations during each generation step.

\subsection{Self-attention based pooler}

Our prompt generator must pool the input instructions of variable lengths to a fixed length. For the pooling operation, the previous literature often chooses average pooling or max pooling \cite{Kim2014ConvolutionalNN,autotrans,zhu2021autonlu}, which are pointed out by the literature \cite{zhu-2021-autorc} that they are prone to weaken important words when the input sequence is long, thus dropping useful information during pooling. Thus, in this work, we utilize the self-attention mechanism in our pooling module $\text{Pooler}()$. Self-Attention assigns each token in the input instruction a weight to indicate the importance of the token. A few crucial tokens to the task will be emphasized, while the less important tokens are ignored. Formally, we initialize a learnable weight matrix $W_{sa} \in \mathbb{R}^{m \times l_{sp}} $, then the self-attention based pooler's calculation processes are:
\begin{align}
\mathbf{U} & = \mathbf{h} W_{sa},  \nonumber \\   
\mathbf{A} & = \text{Softmax}(\mathbf{U}),  \nonumber \\  
\mathbf{p} & = \mathbf{A}^{\intercal} \mathbf{h},
\end{align}
where $\text{Softmax}$ is the softmax function along the first dimension, and $\intercal$ denotes matrix transpose. In the above equations, each column of $W_{sa}$ is a trainable query vector designated to determine the self-attention weights via dot products between this query and each token. Then, the weights are normalized across the sequence dimension via the softmax normalization function. Corresponding to different soft tokens, different query vectors in $W_{sa}$ can aggregate the input instructions in different aspects, thus providing a high-quality summarization of the instruction's semantic information.

\subsection{Learned activation functions}

The previous PEFT literature usually set the activation functions in a PEFT module to be ReLU \cite{Mahabadi2021CompacterEL,pfeiffer-etal-2021-adapterfusion,Liu2022LatePT} and does not discuss whether this setting is optimal. In addition, the PEFT modules' activation functions in different Transformer layers are usually set to be identical. In our initial exploratory experiments (presented in Appendix \ref{sec:appendix_pilot_experiments}), we find that (a) different downstream tasks require different activation functions for the prompt generators in Equation \ref{eq:prompt_generator}. (b) it is beneficial for prompt generators of different depths to have different activation functions. Thus, how can we find an optimal setting for the prompt generators' activation functions? Exhaustive hyper-parameter search is time and GPU-consuming. Thus, we are motivated to set the activation function to be learnable during training.

We resort to rational activation functions to make the activation functions learnable. Empirically introduced as Pad\'{e} Activation Units \cite{Molina2019PadAU}, rational activation functions are learnable and can approximate common activation functions and learn new ones. The rational activation function
$R(x)$ of order $m$, $n$ is defined as follows: 
\begin{equation}
R(x) = \dfrac{ \sum_{j=0}^{m} a_{j} x^{j} }{  1 + \| 
\sum_{i=1}^{n} b_{i} x^{i} \| },   
\end{equation}
where $a_{j}$ and $b_{i}$ are learnable parameters. The rational activation functions are integrated in image classification models \cite{Molina2019PadAU}, sequence modeling \cite{delfosse2021recurrent}, the policy and critic networks in reinforcement learning \cite{Delfosse2021AdaptiveRA}, and Generative Adversarial Networks \cite{Boulle2020RationalNN}.

Inspired by the above literature, we propose learning the activation functions in prompt generators via the rational activation functions when finetuning a downstream task. Denote the set of parameters in the learnable activations as $\Theta$ and the other parameters in the prompt generators as $\Omega$. Following DARTS \cite{Liu2019DARTSDA}, we consider $\Theta$ as architectural parameters and optimize them along with the prompt generators' parameters $\Omega$ via bi-level optimization. Due to limited length, we introduced bi-level optimization to Appendix \ref{sec:appendix_bi_level_opt}.

\subsection{Cross-layer parameter sharing}

To improve our IAPT method's parameter efficiency, we propose sharing the parameters of prompt generators across Transformer layers. Denote the total number of Transformer layers in the LLM as $L$. We ask every $L_s > 0$ prompt generators to (a) share the parameters in $\text{MLP}_{up}$, $\text{MLP}_{down}$, and the learnable activations, (b) but not to share the parameters in the self-attentional $\text{Pooler}$. We will use experiments to show that the self-attention $\text{Pooler}$ is the key that parameters of prompt generators can be shared across layers.

\section{Experiments}

In this section, we conduct a series of experiments and analysis to evaluate our IAPT method.

\subsection{Baselines}

We compare our IAPT framework with the current SOTA PEFT baseline methods. 

\noindent\textbf{Adapter-based tuning} \ We consider the following adapter tuning baselines: (1) Houlsby-Adapter \cite{houlsby2019parameter}; (2) Parallel-Adapter proposed by \citet{He2021TowardsAU}; (3) AdapterDrop \cite{Rckl2020AdapterDropOT}; (4) Learned-Adapter \cite{Zhang2023LearnedAA}.

\noindent\textbf{LoRA and its variants} \ we consider the following LoRA variants as baselines: (a) the original LoRA \cite{hu2021lora}; (b) AdaLoRA \cite{Zhang2023AdaptiveBA}, which adaptively adjust the LoRA ranks among different Transformer modules.

\noindent\textbf{Prompt-based tuning} \ For prompt-based tuning methods, we compare with (a) P-tuning \cite{Liu2021PTuningVP}. (b) P-tuning v2 \cite{Liu2021PTuningVP}. (c) IDPG \cite{Wu2022IDPGAI}. (d) LPT \cite{Liu2022LatePT}. We adjust the tunable parameters of these methods via reparameterization so that their tunable parameters are comparable to our IAPT methods.  

\noindent\textbf{Other PEFT methods} \ We also compare: (a) BitFit \cite{BenZaken2021BitFitSP}, which fine-tunes the model by adding tunable bias terms to the linear layers of LLMs. (b) (IA)$^{3}$ \cite{Liu2022FewShotPF}, which multiplies learnable vectors to the hidden states in different modules of the Transformer layer. (c) SSP \cite{Hu2022SparseSS}, which is a representative work on combining different PEFT methods, including LoRA and BitFit.

The baselines are implemented using their open-sourced codes. We only adjust the hyper-parameters related to tunable parameter numbers to compare the baseline methods and our IAPT method fairly. The hyper-parameter settings for the baselines are detailed in Appendix \ref{sec:appendix_exp_settings}.

\subsection{Datasets and evaluation metrics}

We compare our approach to the baselines on the following benchmark tasks: (a) four benchmark question-answering tasks: SQuAD \cite{rajpurkar-etal-2016-squad} and three tasks from the SuperGLUE benchmark\cite{Wang2019SuperGLUEAS} (BoolQ, COPA, and ReCoRD). (b) three sentence level tasks from GLUE benchmark \cite{Wang2018GLUEAM}, SST-2, RTE, QNLI. (c) a constrained natural language generation task E2E \cite{novikova-etal-2017-e2e}. (d) a mathematical solving dataset GSM8K \cite{Cobbe2021TrainingVT}. (e) a SQL generation task WikiSQL \cite{Zhong2017Seq2SQLGS}. (f) Alpaca dataset \cite{alpaca} for general-purpose instruction tuning, and MT-Bench \cite{2023arXiv230605685Z}, to evaluate the instruction tuning quality of LLMs. The dataset introductions, statistics, and prompt-response templates for the above tasks are detailed in Appendix \ref{sec:appendix_datasets}. The above tasks' evaluation metrics or protocols are in Appendix \ref{sec:appendix_evaluations}.

\begin{table*}[tb!]
\centering
\resizebox{0.96\textwidth}{!}{
\begin{tabular}{c|c|ccccccc}
\hline
\multirow{2}*{\textbf{Method}}   &   \textbf{Tunable}   &     \textbf{SST-2}   &    \textbf{RTE}   &   \textbf{QNLI}   &   \textbf{BoolQ}  &  \textbf{COPA}    &   \textbf{ReCoRD}   &    \textbf{SQuAD}    \\ 

&  \textbf{Params}  &   \textbf{(acc)}   &   \textbf{(acc)}     &  \textbf{(acc)}   &   \textbf{(acc)}  &   \textbf{(acc)}  &   \textbf{(f1-em)}   &   \textbf{(f1-em)}     \\
\hline

\multicolumn{9}{c}{\textbf{\emph{Baselines}}}  \\
\hline

Housbly-Adapter   &    9.4M    &  92.9   &  80.6  &  92.4   &  84.5  &  90.4    &  89.8    &  87.3      \\

Parallel-Adapters  &    9.4M   &   93.0   &   80.5   &  92.5   & 85.1   &  90.2   &  90.1    &  87.7     \\

AdapterDrop  &    9.2M  &    92.7  &   80.1   &  92.3   &  84.5   &  89.8   &  89.8   &  87.4        \\

Learned-Adapter   &   9.5M   &    93.6    &  81.5   &  92.4    &  86.2   &  90.4   &  90.1   &  87.6       \\

\hdashline

LoRA   &     10.0M   &   93.6   &   82.6    &  92.5   &   \underline{86.7}    &   90.7 &    90.2    &   \underline{87.7}    \\
 
AdaLoRA   &  10.0M   &   \underline{93.6}   &  \underline{82.9}   &  92.6    &  86.6   &  90.8  &  \underline{90.5}  &  87.5     \\


\hdashline
BitFit &   10.9M   &    92.9    &   81.9    &  92.2 
  &   85.6    &  90.5   &  89.8   &   87.2    \\

(IA)$^{3}$  &    9.8M   &      93.0    &    82.7  &  92.5  &  86.4   &   90.7   &   90.1   &   87.6         \\

SSP &   8.6M   &   93.5   & 82.6  &   \underline{92.6}   &  86.4   &     \underline{91.1}   &  90.0   &  87.4    \\

\hdashline

P-Tuning  &    9.4M   &   92.4  &  79.7    &  91.9  &   84.1   &   89.6   &    89.2    &   86.5      \\

P-tuning v2    &    9.4M    &    92.8    &   80.6  &  92.1   &   85.2   &  90.1   &   89.4   &  86.9       \\

IDPG   &    8.4M     &   92.6   &  80.8   &  92.2   &  85.3   &   90.1   &  89.6   &   87.2    \\

LPT  &    8.4M   &    92.8    &   81.3   &  92.3   &   85.7   &   90.2   &   89.9   &  87.4     \\

\hdashline

\hline
\multicolumn{9}{c}{\textbf{\emph{Our proposed methods}}}  \\
\hline

IAPT   &   8.4M   &   \textbf{94.3}  &  \textbf{83.9}   &  \textbf{93.2}  &  \textbf{87.5}  &  \textbf{91.9}   &   \textbf{91.2}    &   \textbf{88.5}      \\

\hline
\end{tabular}}

\caption{\label{tab:results_main_1} The Overall comparison of the three GLUE tasks and four question-answering tasks. The backbone model is LlaMA-2 7B. We report the median performance over five random seeds. Bold and Underline indicate the best and the second-best results. The metric for each task is explained in Appendix \ref{sec:appendix_evaluations}.} 
\end{table*}

\subsection{Experiment Settings}
\label{subsec:experimental_settings}

\noindent\textbf{Computing infrastures} \quad We run all our experiments on NVIDIA A40 (48GB) GPUs. 

\noindent\textbf{Pretrained backbones} \quad The main experiments use the most recent open-sourced LLM, LlaMA-2 7B released by Meta \cite{Touvron2023Llama2O} as the pretrained backbone model. In the ablation studies, we will also use the GPT2-large model \cite{radford2019language} and Pythia-1.4B \cite{Biderman2023PythiaAS}. 

\noindent\textbf{Prediction heads} \quad When fine-tuning LlaMA-2 7B, we only consider the supervised fine-tuning (SFT) setting \cite{ouyang2022training}. After receiving a prompt or instruction, all the predictions are generated using the language modeling head (LM head). No additional prediction heads are installed for making categorical or numerical predictions. For decoding during inference, we use beam search with beam size 3.

\noindent\textbf{Hyper-parameters for the IAPT framework} \quad In our experiments, unless otherwise specified, we set: (a) the length of soft prompts is $l_{sp}=4$, (b) the bottleneck dimension $m$ of the IAPT prompt generator is 256, (c) every $L_{s} = 4$ layers share the prompt generators' parameters except for the self-attention poolers, and (d) the hyper-parameters of the rational activation are $m = 6$, $n=5$, and the learnable parameters $a_j$ and $b_i$ are initialized by approximating the GeLU activation function. Under the above settings, our IAPT method will introduce 8.4M tunable parameters to the LlaMA-2 7B backbone. The hyper-parameters for training are specified in Appendix \ref{sec:appendix_exp_settings}.











\noindent\textbf{Reproducibility} \quad We run each task under five different random seeds and report the median performance on the test set of each task. 

Due to limited length, other experimental settings for the baseline methods and the training procedure are in Appendix \ref{sec:appendix_exp_settings}.

\subsection{Main results}
\label{subsec:main_results}

\noindent \textbf{Results on the GLUE and SuperGLUE tasks} \quad The experimental results on the three classification tasks and 4 question answering tasks are presented in Table \ref{tab:results_main_1}. We present the number of tunable parameters in the second column of Table \ref{tab:results_main_1}. Table \ref{tab:results_main_1} reveals that our IAPT method outperforms the baseline methods across all seven tasks, with comparable or fewer tunable parameters. In particular, IAPT outperforms the previous SOTA prompt tuning methods like P-tuning V2 and LPT and the strong LoRA style baselines like LoRA and AdaLoRA with comparable parameters. These results demonstrate that our method is good at downstream task adaptation of large language models.  


\noindent \textbf{Results on the three specialized generation task} \quad For the E2E, GSM8K, and WikiSQL benchmarks, the results are reported in Table \ref{tab:results_nlg}. The results show that our IAPT method outperforms LoRA, AdaLoRA, and LPT on the three tasks.

\begin{table}[tb!]
\centering
\resizebox{0.45\textwidth}{!}{
\begin{tabular}{c|ccc}
\hline
\multirow{2}*{\textbf{Method}}   &   \textbf{E2E}    &   
  \textbf{GSM8K}    &    \textbf{WikiSQL}     \\ 

&  \textbf{(rouge-l)}  &   \textbf{(acc)}    &  \textbf{(acc)}   \\
\hline

\hline

LPT    &    70.4    &     34.2   &    84.3    \\
LoRA   &   70.7    &     35.1    &    85.4    \\

AdaLoRA   &  70.8    &   35.2    &   85.2  \\
\hdashline
IAPT   &     71.3   &   36.4     &    85.9   \\

\hline
\end{tabular}}
\caption{\label{tab:results_nlg} Results for different PEFT methods on the E2E, GSM8K, and WikiSQL benchmark. The backbone LM is LlaMA-2 7B. }
\end{table}

\begin{table}[tb!]
\centering
\resizebox{0.46\textwidth}{!}{
\begin{tabular}{c|cc}
\hline
\textbf{Method}   &    \textbf{Avg GPT-4 score} ($\uparrow$)   &    \textbf{ ROUGE-L} ($\uparrow$)    \\ 
\hline
AdaLoRA    &   6.95   &    51.1 \\
\hdashline
IAPT   &   7.19   &   52.8  \\
\hline

\end{tabular}}
\caption{\label{tab:results_alpaca} The performance of instruction tuning using the AdaLoRA and IAPT methods. The backbone model is LlaMA-2 7B. $\uparrow$ means the metric is higher the better. }
\end{table}

\begin{figure*}[ht]	
\centering
\subfigure[BoolQ]{%
\includegraphics[width=0.46\textwidth]{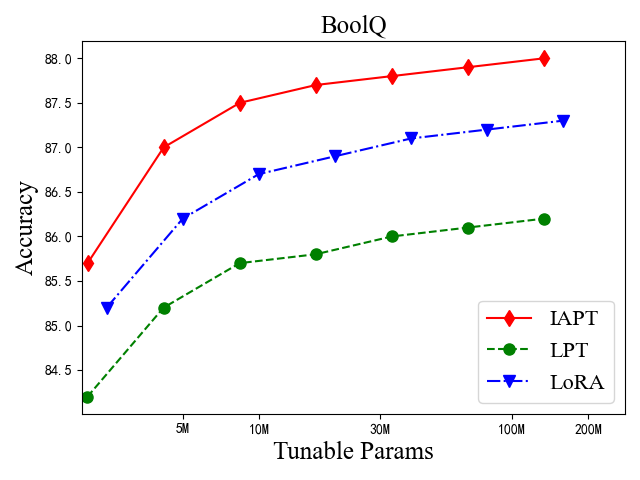}
\label{subfig:different_rank_boolq}
}
\subfigure[E2E]{%
\includegraphics[width=0.46\textwidth]{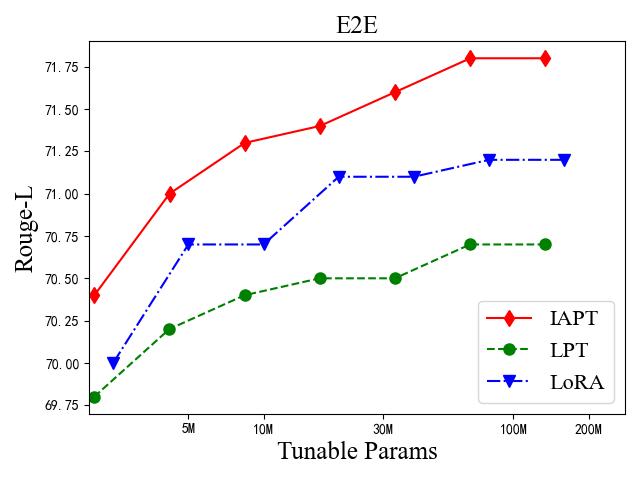}
\label{subfig:different_rank_e2e}
}
\caption{Performances under different tunable parameter budgets. The $x$-axis represents the number of tunable parameters, and the $y$-axis represents the performance score. }
\label{fig:different_tunable_params}
\end{figure*}

\noindent \textbf{Results for general-purpose instruction tuning} \quad After the LlaMA-2 7B is fine-tuned on the Alpaca dataset with our IAPT and AdaLoRA methods, we utilize the 80 instructions in the MT-Bench as the test set. We follow the current standard practice of utilizing GPT-4 as an unbiased reviewer \cite{2023arXiv230605685Z}. The protocol of utilizing GPT-4 as the reviewer and scorer is specified in Appendix \ref{sec:appendix_evaluations}. The average score provided by GPT-4 is presented in Table \ref{tab:results_alpaca}, along with the ROUGE-L scores calculated by considering the GPT-4's answers as ground truth. Consistent with the previous experiments (Table \ref{tab:results_main_1} and \ref{tab:results_nlg}), our IAPT method outperforms the AdaLoRA method in terms of the GPT-4 evaluation scores and ROUGE-L, demonstrating that IAPT can enhance the instruction tuning quality of large language models. A case study of answers generated by different methods is presented in Table \ref{tab:results_alpaca_examples_1} of Appendix \ref{sec:appendix_case_study}, showcasing that IAPT leads to better instruction-tuned LLMs.

\subsection{Ablation studies and analysis}

\noindent\textbf{Analysis of the inference efficiency} \quad To demonstrate the inference efficiency of our IAPT method, we now compare the GPU memory and generation speed of IAPT and LoRA. In this experiment, LoRA parameters are not merged to the backbone to mimic the single-LLM multi-tenant setting \cite{Chen2023PunicaML}. The detailed settings for efficiency analysis are presented in Appendix \ref{sec:instruction_for_efficiency_analysis}. We present two metrics for measuring efficiency: (a) peak memory cost during generation. (b) tokens generated per second (tps). The results are presented in Table \ref{tab:results_efficiency_analysis}.

\begin{table}[tb!]
\centering
\resizebox{0.46\textwidth}{!}{
\begin{tabular}{c|ccc}
\hline
\textbf{Method}   &    \textbf{Beam size}  &  \textbf{Speed (tps)}   &   \textbf{Memory cost (MiB)}     \\ 
\hline

\multirow{ 2}{*}{ LoRA }   &   1    &   25.1    &   14616   \\
    &   3   &    21.9    &    16104  \\

\hdashline
\multirow{ 2}{*}{ IAPT }   &   1   &   34.4   &   14490  \\
    &   3   &  27.9   &   15946  \\
    
\hline
\end{tabular}}
\caption{\label{tab:results_efficiency_analysis} The memory and speed of LlaMA-2 7B for generating responses given the input instruction (Appendix \ref{sec:instruction_for_efficiency_analysis}), with different PEFT methods.  }
\end{table}


























From Table \ref{tab:results_efficiency_analysis}, one can see that under beam sizes 1 and 3, the IAPT method has a comparable memory cost with LoRA, but the generation speed is significantly higher. The speed advantages of IAPT come from the following factors: (a) our method only adds four soft tokens, which is relatively short compared to the instructions received by the modern LLMs \cite{ouyang2022training}. (b) our prompt generators are lightweight and efficient during inference. (c) The soft prompts are only generated once when an input instruction is passed to the LLM and right before generating the first new token. The soft prompts are integrated into the KV cache in the following generation steps. In contrast, the LoRA method requires the model to call the LoRA modules at each generation step, resulting in higher latency.

\noindent\textbf{Visualization of the learned activation functions} \quad In Figure \ref{fig:activation_funct} of Appendix \ref{sec:appendix_visualize_activation}, we visualize the learned activation functions of the prompt generators for every $L_{s} = 4$ Transformer layers after fine-tuning on the Alpaca dataset. Rational GeLU is the rational function approximating the GeLU activation and is used to initialize the learnable activation functions for the prompt generators. Rational GeLU and GeLU are overlapping with each other. As shown in Figure \ref{fig:activation_funct}, we can see that (a) the learned activation function differs from the GeLU activation function but still has a similar shape to GeLU. (b) The learned activation functions are different across different Transformer layers. We can see that the learned activations adapted to the fine-tuning dataset and can extract suitable features for providing suitable soft prompts.

\noindent\textbf{Ablation study of IAPT framework} \quad We now consider the following variants of IAPT: (a) IAPT-1 substitutes the self-attention pooler to average pooling. (b) IAPT-2 sets $L_{s} = 16$ and $m = 1024$. (c) IAPT-3 sets $L_{s} = 1$ and $m = 64$. (d) IAPT-4 uses the GeLU activation function for every prompt generator. (e) IAPT-5 uses the ReLU activation function for every prompt generator. (f) IAPT-6 uses ReLU for the first 16 layers' prompt generators, and GeLU for the deeper 16 layers'. (h) IAPT-7 uses GeLU for the first 16 layers' prompt generators, and ReLU for the deeper 16 layers'. The experimental results on the BoolQ, E2E, and SQuAD tasks are reported in Table \ref{tab:appendix_ablation}. The results show that IAPT under the default settings (as in Table \ref{tab:results_main_1}) outperforms the four variants. In addition, (a) comparing IAPT-1 to IAPT shows that the self-attention poolers provide more practical information aggregation. In addition, self-attention poolers provide adaptive feature extraction for cross-layer parameter sharing. (b) Comparing IAPT to IAPT-2 and IAPT-3 demonstrates that under the comparable tunable parameters, cross-layer parameter sharing of prompt generators allows for higher values of $m$, thus improving the capacity of IAPT. However, aggressively sharing prompt generators across layers could hurt downstream performance. (c) Comparing IAPT to IAPT-4, IAPT-5, IAPT-6 and IAPT-7 demonstrates the necessity of learning activation functions for the prompt generators.

\begin{table}[tb!]
\centering
\resizebox{0.42\textwidth}{!}{
\begin{tabular}{c|ccc}
\hline
\multirow{2}*{\textbf{Method}}    &     \textbf{BoolQ}     &   \textbf{E2E}   &    \textbf{SQuAD}  \\ 

&    \textbf{(acc)}  &   \textbf{(rouge-l)}   &   \textbf{(f1-em)}  \\
\hline
IAPT   &     \textbf{87.5}    &    \textbf{71.3}   &     \textbf{88.5}       \\
\hdashline

IAPT-1   &     86.9   &   70.7   &  88.0    \\
IAPT-2   &     86.2   &   70.2   &  87.3   \\
IAPT-3   &     87.2   &   71.0   &  88.1  \\
IAPT-4   &   86.7   &   70.6   &   87.8     \\
IAPT-5   &    86.4  &  70.7  &  87.7       \\
IAPT-6   &    86.8    &  70.7  &  88.0      \\
IAPT-7   &    86.6   &  70.5   &  87.9  \\
\hline
\end{tabular}}

\caption{\label{tab:appendix_ablation} The comparison of IAPT's variants on the BoolQ, E2E, and SQuAD tasks. The backbone model is LlaMA-2 7B. } 
\end{table}

\noindent\textbf{Comparisons under different budgets of tunable parameters} \quad We vary the budget of tunable parameters for IAPT by modifying the values of $m=256$ to \{64, 128, 512, 1024, 2048, 4096\}. We also vary the LPT and LoRA methods' tunable parameter numbers. The experimental results on the BoolQ and E2E tasks are presented in Figure \ref{subfig:different_rank_boolq} and \ref{subfig:different_rank_e2e}. The results show that under different tunable parameter budgets, our IAPT method can consistently outperform the LoRA and LPT methods.

\noindent\textbf{Effects of different lengths of soft prompts} \quad We vary the length $l_{sp}$ from 4 to \{1, 2, 8, 16, 32\} for IAPT and LPT. The experimental results on the BoolQ task are presented in Figure \ref{fig:BoolQ_different_length} of Appendix \ref{sec:appendix_ablation_prompt_length}. The results show that our IAPT method is less sensitive to the prompt length in terms of downstream performance and performs better than the LPT baseline under different prompt lengths. IAPT effectively aggregates the semantic features of the input instructions with the help of the self-attention pooler and learnable activations, thus obtaining better downstream performances.

\noindent\textbf{Ablation on the pretrained backbones} \quad Our main experiments are conducted on the LlaMA-2 7B model. To demonstrate the broad applicability of our method, we now conduct experiments on GPT2-large and Pythia-1.4b. The results are reported in Table \ref{tab:results_different_backbones}. We can see that on these two backbones, our method can also outperform the baseline methods.

\section{Conclusion}

This work presents the instruction aware prompt tuning (IAPT) method, an innovative method for the parameter-efficient fine-tuning of large language models. Upon the hypothesis that different input instructions require different soft prompts, we propose to generate soft prompts from the input instructions. We propose three recipes for improving our framework's downstream performance: (a) self-attention pooling; (b) learning different activation functions during fine-tuning for different prompt generators of different depth; (c) cross-layer parameter sharing of prompt generators. Our method is convenient to implement and off-the-shelf. Experiments on various tasks demonstrate that our IAPT method outperforms the baseline methods, while being efficient for inference.

\section*{Limitations}

We showed that our proposed method can greatly improve the performance of parameter-efficient tuning on diverse tasks and different pretrained models (i.e., LlaMA-2 7B, RoBERTa-large and GPT2-large). However, we acknowledge the following limitations: (a) the more super-sized open-sourced LLMs, such as LlaMA-2 13B and 70B, are not experimented due to limited computation resources. (b) Other tasks in natural language processing, like information extraction, were also not considered. But our framework can be easily transferred to other backbone architectures and different types of tasks. It would be of interest to investigate if the superiority of our method holds for other large-scaled backbone models and other types of tasks. And we will explore it in future work.

\section*{Ethics Statement}

The finding and proposed method aims to improve the soft prompt based tuning in terms of better downstream performances whiling pursuing efficiency. The used datasets are widely used in previous work and, to our knowledge, do not have any attached privacy or ethical issues. In this work, we have experimented with LlaMA-2 7B, a modern large language model. As with all LLMs, LlaMA-2’s potential outputs cannot be predicted in advance, and the model may in some instances produce inaccurate, biased or other objectionable responses to user prompts. However, this work's intent is to conduct research on different fine-tuning methods for LLMs, not building applications to general users. In the future, we would like to conduct further tests to see how our method affects the safety aspects of LLMs.

\bibliography{custom}

\begin{thebibliography}{106}
\expandafter\ifx\csname natexlab\endcsname\relax\def\natexlab#1{#1}\fi

\bibitem[{Aghajanyan et~al.(2021)Aghajanyan, Gupta, and Zettlemoyer}]{aghajanyan-etal-2021-intrinsic}
Armen Aghajanyan, Sonal Gupta, and Luke Zettlemoyer. 2021.
\newblock \href {https://doi.org/10.18653/v1/2021.acl-long.568} {Intrinsic dimensionality explains the effectiveness of language model fine-tuning}.
\newblock In \emph{Proceedings of the 59th Annual Meeting of the Association for Computational Linguistics and the 11th International Joint Conference on Natural Language Processing (Volume 1: Long Papers)}, pages 7319--7328, Online. Association for Computational Linguistics.

\bibitem[{Ben-Zaken et~al.(2021)Ben-Zaken, Ravfogel, and Goldberg}]{BenZaken2021BitFitSP}
Elad Ben-Zaken, Shauli Ravfogel, and Yoav Goldberg. 2021.
\newblock Bitfit: Simple parameter-efficient fine-tuning for transformer-based masked language-models.
\newblock \emph{ArXiv}, abs/2106.10199.

\bibitem[{Biderman et~al.(2023)Biderman, Schoelkopf, Anthony, Bradley, O'Brien, Hallahan, Khan, Purohit, Prashanth, Raff, Skowron, Sutawika, and van~der Wal}]{Biderman2023PythiaAS}
Stella Biderman, Hailey Schoelkopf, Quentin~G. Anthony, Herbie Bradley, Kyle O'Brien, Eric Hallahan, Mohammad~Aflah Khan, Shivanshu Purohit, USVSN~Sai Prashanth, Edward Raff, Aviya Skowron, Lintang Sutawika, and Oskar van~der Wal. 2023.
\newblock Pythia: A suite for analyzing large language models across training and scaling.
\newblock \emph{ArXiv}, abs/2304.01373.

\bibitem[{Boull'e et~al.(2020)Boull'e, Nakatsukasa, and Townsend}]{Boulle2020RationalNN}
Nicolas Boull'e, Yuji Nakatsukasa, and Alex Townsend. 2020.
\newblock Rational neural networks.
\newblock \emph{ArXiv}, abs/2004.01902.

\bibitem[{Chen et~al.(2023)Chen, Ye, Wu, Zhuo, Ceze, of~Washington, and University}]{Chen2023PunicaML}
Lequn Chen, Zihao Ye, Yongji Wu, Danyang Zhuo, Luis Ceze, Arvind Krishnamurthy~University of~Washington, and Duke University. 2023.
\newblock Punica: Multi-tenant lora serving.
\newblock \emph{ArXiv}, abs/2310.18547.

\bibitem[{Cobbe et~al.(2021)Cobbe, Kosaraju, Bavarian, Chen, Jun, Kaiser, Plappert, Tworek, Hilton, Nakano, Hesse, and Schulman}]{Cobbe2021TrainingVT}
Karl Cobbe, Vineet Kosaraju, Mohammad Bavarian, Mark Chen, Heewoo Jun, Lukasz Kaiser, Matthias Plappert, Jerry Tworek, Jacob Hilton, Reiichiro Nakano, Christopher Hesse, and John Schulman. 2021.
\newblock Training verifiers to solve math word problems.
\newblock \emph{ArXiv}, abs/2110.14168.

\bibitem[{Cui et~al.(2023)Cui, Yuan, Ding, Yao, Zhu, Ni, Xie, Liu, and Sun}]{Cui2023UltraFeedbackBL}
Ganqu Cui, Lifan Yuan, Ning Ding, Guanming Yao, Wei Zhu, Yuan Ni, Guotong Xie, Zhiyuan Liu, and Maosong Sun. 2023.
\newblock \href {https://api.semanticscholar.org/CorpusID:263605623} {Ultrafeedback: Boosting language models with high-quality feedback}.
\newblock \emph{ArXiv}, abs/2310.01377.

\bibitem[{Delfosse et~al.(2021)Delfosse, Schramowski, Molina, and Kersting}]{delfosse2021recurrent}
Quentin Delfosse, Patrick Schramowski, Alejandro Molina, and Kristian Kersting. 2021.
\newblock Recurrent rational networks.
\newblock \emph{arXiv preprint arXiv:2102.09407}.

\bibitem[{Delfosse et~al.(2023)Delfosse, Schramowski, Mundt, Molina, and Kersting}]{Delfosse2021AdaptiveRA}
Quentin Delfosse, Patrick Schramowski, Martin Mundt, Alejandro Molina, and Kristian Kersting. 2023.
\newblock Adaptive rational activations to boost deep reinforcement learning.
\newblock In \emph{The Twelfth International Conference on Learning Representations}.

\bibitem[{{Dettmers} et~al.(2023){Dettmers}, {Pagnoni}, {Holtzman}, and {Zettlemoyer}}]{2023arXiv230514314D}
Tim {Dettmers}, Artidoro {Pagnoni}, Ari {Holtzman}, and Luke {Zettlemoyer}. 2023.
\newblock \href {https://doi.org/10.48550/arXiv.2305.14314} {{QLoRA: Efficient Finetuning of Quantized LLMs}}.
\newblock \emph{arXiv e-prints}, page arXiv:2305.14314.

\bibitem[{Ding et~al.(2023)Ding, Lv, Wang, Chen, Zhou, Liu, and Sun}]{Ding2023SparseLA}
Ning Ding, Xingtai Lv, Qiaosen Wang, Yulin Chen, Bowen Zhou, Zhiyuan Liu, and Maosong Sun. 2023.
\newblock \href {https://api.semanticscholar.org/CorpusID:265294736} {Sparse low-rank adaptation of pre-trained language models}.
\newblock In \emph{Conference on Empirical Methods in Natural Language Processing}.

\bibitem[{Ding et~al.(2022)Ding, Qin, Yang, Wei, Yang, Su, Hu, Chen, Chan, Chen, Yi, Zhao, Wang, Liu, Zheng, Chen, Liu, Tang, Li, and Sun}]{Ding2022DeltaTA}
Ning Ding, Yujia Qin, Guang Yang, Fu~Wei, Zonghan Yang, Yusheng Su, Shengding Hu, Yulin Chen, Chi-Min Chan, Weize Chen, Jing Yi, Weilin Zhao, Xiaozhi Wang, Zhiyuan Liu, Haitao Zheng, Jianfei Chen, Yang Liu, Jie Tang, Juan Li, and Maosong Sun. 2022.
\newblock Delta tuning: A comprehensive study of parameter efficient methods for pre-trained language models.
\newblock \emph{ArXiv}, abs/2203.06904.

\bibitem[{Gao et~al.(2023)Gao, Zhu, Gao, and Yin}]{Gao2023FPABEEFE}
Xiangxiang Gao, Wei Zhu, Jiasheng Gao, and Congrui Yin. 2023.
\newblock \href {https://api.semanticscholar.org/CorpusID:258546016} {F-pabee: Flexible-patience-based early exiting for single-label and multi-label text classification tasks}.
\newblock \emph{ArXiv}, abs/2305.11916.

\bibitem[{Guo et~al.(2021{\natexlab{a}})Guo, Rush, and Kim}]{guo-etal-2021-parameter}
Demi Guo, Alexander Rush, and Yoon Kim. 2021{\natexlab{a}}.
\newblock \href {https://doi.org/10.18653/v1/2021.acl-long.378} {Parameter-efficient transfer learning with diff pruning}.
\newblock In \emph{Proceedings of the 59th Annual Meeting of the Association for Computational Linguistics and the 11th International Joint Conference on Natural Language Processing (Volume 1: Long Papers)}, pages 4884--4896, Online. Association for Computational Linguistics.

\bibitem[{Guo et~al.(2021{\natexlab{b}})Guo, Ni, Wang, Zhu, and Xie}]{Guo2021GlobalAD}
Zhao Guo, Yuan Ni, Keqiang Wang, Wei Zhu, and Guo~Tong Xie. 2021{\natexlab{b}}.
\newblock \href {https://api.semanticscholar.org/CorpusID:236477700} {Global attention decoder for chinese spelling error correction}.
\newblock In \emph{Findings}.

\bibitem[{Guo et~al.(2021{\natexlab{c}})Guo, Ni, Wang, Zhu, and Xie}]{guo-etal-2021-global}
Zhao Guo, Yuan Ni, Keqiang Wang, Wei Zhu, and Guotong Xie. 2021{\natexlab{c}}.
\newblock \href {https://doi.org/10.18653/v1/2021.findings-acl.122} {Global attention decoder for {C}hinese spelling error correction}.
\newblock In \emph{Findings of the Association for Computational Linguistics: ACL-IJCNLP 2021}, pages 1419--1428, Online. Association for Computational Linguistics.

\bibitem[{He et~al.(2021)He, Zhou, Ma, Berg-Kirkpatrick, and Neubig}]{He2021TowardsAU}
Junxian He, Chunting Zhou, Xuezhe Ma, Taylor Berg-Kirkpatrick, and Graham Neubig. 2021.
\newblock Towards a unified view of parameter-efficient transfer learning.
\newblock \emph{ArXiv}, abs/2110.04366.

\bibitem[{Houlsby et~al.(2019)Houlsby, Giurgiu, Jastrzebski, Morrone, De~Laroussilhe, Gesmundo, Attariyan, and Gelly}]{houlsby2019parameter}
Neil Houlsby, Andrei Giurgiu, Stanislaw Jastrzebski, Bruna Morrone, Quentin De~Laroussilhe, Andrea Gesmundo, Mona Attariyan, and Sylvain Gelly. 2019.
\newblock Parameter-efficient transfer learning for nlp.
\newblock In \emph{International Conference on Machine Learning}, pages 2790--2799. PMLR.

\bibitem[{Hu et~al.(2021)Hu, Shen, Wallis, Allen-Zhu, Li, Wang, Wang, and Chen}]{hu2021lora}
Edward~J Hu, Yelong Shen, Phillip Wallis, Zeyuan Allen-Zhu, Yuanzhi Li, Shean Wang, Lu~Wang, and Weizhu Chen. 2021.
\newblock Lora: Low-rank adaptation of large language models.
\newblock \emph{arXiv preprint arXiv:2106.09685}.

\bibitem[{Hu et~al.(2022)Hu, Zhang, Ding, Wang, Wang, Liu, and Sun}]{Hu2022SparseSS}
Shengding Hu, Zhen Zhang, Ning Ding, Yadao Wang, Yasheng Wang, Zhiyuan Liu, and Maosong Sun. 2022.
\newblock Sparse structure search for parameter-efficient tuning.
\newblock \emph{ArXiv}, abs/2206.07382.

\bibitem[{Huang et~al.(2023)Huang, Bai, Zhu, Zhang, Zhang, Su, Liu, Lv, Zhang, Lei et~al.}]{huang2023c}
Yuzhen Huang, Yuzhuo Bai, Zhihao Zhu, Junlei Zhang, Jinghan Zhang, Tangjun Su, Junteng Liu, Chuancheng Lv, Yikai Zhang, Jiayi Lei, et~al. 2023.
\newblock C-eval: A multi-level multi-discipline chinese evaluation suite for foundation models.
\newblock \emph{arXiv preprint arXiv:2305.08322}.

\bibitem[{Jie and Deng(2022)}]{Jie2022ConvolutionalBA}
Shibo Jie and Zhifang Deng. 2022.
\newblock Convolutional bypasses are better vision transformer adapters.
\newblock \emph{ArXiv}, abs/2207.07039.

\bibitem[{Kim(2014)}]{Kim2014ConvolutionalNN}
Yoon Kim. 2014.
\newblock \href {https://doi.org/10.3115/v1/D14-1181} {Convolutional neural networks for sentence classification}.
\newblock In \emph{Proceedings of the 2014 Conference on Empirical Methods in Natural Language Processing ({EMNLP})}, pages 1746--1751, Doha, Qatar. Association for Computational Linguistics.

\bibitem[{Kopiczko et~al.(2023)Kopiczko, Blankevoort, and Asano}]{Kopiczko2023VeRAVR}
Dawid~Jan Kopiczko, Tijmen Blankevoort, and Yuki~Markus Asano. 2023.
\newblock \href {https://api.semanticscholar.org/CorpusID:264172315} {Vera: Vector-based random matrix adaptation}.
\newblock \emph{ArXiv}, abs/2310.11454.

\bibitem[{Le et~al.(2021)Le, Bertolini, No'e, and Clevert}]{Le2021ParameterizedHG}
Tuan Le, Marco Bertolini, Frank No'e, and Djork-Arn{\'e} Clevert. 2021.
\newblock Parameterized hypercomplex graph neural networks for graph classification.
\newblock In \emph{International Conference on Artificial Neural Networks}.

\bibitem[{Lester et~al.(2021)Lester, Al-Rfou, and Constant}]{lester2021power}
Brian Lester, Rami Al-Rfou, and Noah Constant. 2021.
\newblock The power of scale for parameter-efficient prompt tuning.
\newblock \emph{arXiv preprint arXiv:2104.08691}.

\bibitem[{Li et~al.(2023{\natexlab{a}})Li, Zhang, Koto, Yang, Zhao, Gong, Duan, and Baldwin}]{li2023cmmlu}
Haonan Li, Yixuan Zhang, Fajri Koto, Yifei Yang, Hai Zhao, Yeyun Gong, Nan Duan, and Timothy Baldwin. 2023{\natexlab{a}}.
\newblock Cmmlu: Measuring massive multitask language understanding in chinese.
\newblock \emph{arXiv preprint arXiv:2306.09212}.

\bibitem[{Li and Liang(2021)}]{li2021prefix}
Xiang~Lisa Li and Percy Liang. 2021.
\newblock Prefix-tuning: Optimizing continuous prompts for generation.
\newblock \emph{arXiv preprint arXiv:2101.00190}.

\bibitem[{Li et~al.(2023{\natexlab{b}})Li, Lv, Yan, Lin, Zhu, Ni, Xie, Wang, and Qiu}]{Li2023UnifiedDR}
Xiaonan Li, Kai Lv, Hang Yan, Tianya Lin, Wei Zhu, Yuan Ni, Guo~Tong Xie, Xiaoling Wang, and Xipeng Qiu. 2023{\natexlab{b}}.
\newblock \href {https://api.semanticscholar.org/CorpusID:258557751} {Unified demonstration retriever for in-context learning}.
\newblock \emph{ArXiv}, abs/2305.04320.

\bibitem[{Li et~al.(2019)Li, Zhang, Zhu, Li, Ni, Gao, Yan, and Xie}]{li-etal-2019-pingan}
Xiepeng Li, Zhexi Zhang, Wei Zhu, Zheng Li, Yuan Ni, Peng Gao, Junchi Yan, and Guotong Xie. 2019.
\newblock \href {https://doi.org/10.18653/v1/D19-6011} {Pingan smart health and {SJTU} at {COIN} - shared task: utilizing pre-trained language models and common-sense knowledge in machine reading tasks}.
\newblock In \emph{Proceedings of the First Workshop on Commonsense Inference in Natural Language Processing}, pages 93--98, Hong Kong, China. Association for Computational Linguistics.

\bibitem[{Liu et~al.(2019)Liu, Simonyan, and Yang}]{Liu2019DARTSDA}
Hanxiao Liu, Karen Simonyan, and Yiming Yang. 2019.
\newblock Darts: Differentiable architecture search.
\newblock \emph{ArXiv}, abs/1806.09055.

\bibitem[{Liu et~al.(2022{\natexlab{a}})Liu, Tam, Muqeeth, Mohta, Huang, Bansal, and Raffel}]{Liu2022FewShotPF}
Haokun Liu, Derek Tam, Mohammed Muqeeth, Jay Mohta, Tenghao Huang, Mohit Bansal, and Colin~A Raffel. 2022{\natexlab{a}}.
\newblock Few-shot parameter-efficient fine-tuning is better and cheaper than in-context learning.
\newblock \emph{Advances in Neural Information Processing Systems}, 35:1950--1965.

\bibitem[{Liu et~al.(2022{\natexlab{b}})Liu, Sun, Huang, and Qiu}]{Liu2022LatePT}
Xiangyang Liu, Tianxiang Sun, Xuanjing Huang, and Xipeng Qiu. 2022{\natexlab{b}}.
\newblock Late prompt tuning: A late prompt could be better than many prompts.
\newblock \emph{ArXiv}, abs/2210.11292.

\bibitem[{Liu et~al.(2021)Liu, Ji, Fu, Du, Yang, and Tang}]{Liu2021PTuningVP}
Xiao Liu, Kaixuan Ji, Yicheng Fu, Zhengxiao Du, Zhilin Yang, and Jie Tang. 2021.
\newblock P-tuning v2: Prompt tuning can be comparable to fine-tuning universally across scales and tasks.
\newblock \emph{ArXiv}, abs/2110.07602.

\bibitem[{Liu et~al.(2022{\natexlab{c}})Liu, Ji, Fu, Tam, Du, Yang, and Tang}]{Liu2022PTuningPT}
Xiao Liu, Kaixuan Ji, Yicheng Fu, Weng~Lam Tam, Zhengxiao Du, Zhilin Yang, and Jie Tang. 2022{\natexlab{c}}.
\newblock P-tuning: Prompt tuning can be comparable to fine-tuning across scales and tasks.
\newblock In \emph{Annual Meeting of the Association for Computational Linguistics}.

\bibitem[{Liu et~al.(2024)Liu, Lyn, Zhu, Tian, and Graham}]{liu2024alora}
Zequan Liu, Jiawen Lyn, Wei Zhu, Xing Tian, and Yvette Graham. 2024.
\newblock Alora: Allocating low-rank adaptation for fine-tuning large language models.
\newblock \emph{arXiv preprint arXiv:2403.16187}.

\bibitem[{Mahabadi et~al.(2021)Mahabadi, Henderson, and Ruder}]{Mahabadi2021CompacterEL}
Rabeeh~Karimi Mahabadi, James Henderson, and Sebastian Ruder. 2021.
\newblock Compacter: Efficient low-rank hypercomplex adapter layers.
\newblock In \emph{NeurIPS}.

\bibitem[{Mangrulkar et~al.(2022)Mangrulkar, Gugger, Debut, Belkada, Paul, and Bossan}]{peft}
Sourab Mangrulkar, Sylvain Gugger, Lysandre Debut, Younes Belkada, Sayak Paul, and Benjamin Bossan. 2022.
\newblock Peft: State-of-the-art parameter-efficient fine-tuning methods.
\newblock \url{https://github.com/huggingface/peft}.

\bibitem[{Mishra et~al.(2021)Mishra, Khashabi, Baral, and Hajishirzi}]{Mishra2021CrossTaskGV}
Swaroop Mishra, Daniel Khashabi, Chitta Baral, and Hannaneh Hajishirzi. 2021.
\newblock \href {https://api.semanticscholar.org/CorpusID:237421373} {Cross-task generalization via natural language crowdsourcing instructions}.
\newblock In \emph{Annual Meeting of the Association for Computational Linguistics}.

\bibitem[{Molina et~al.(2019)Molina, Schramowski, and Kersting}]{Molina2019PadAU}
Alejandro Molina, Patrick Schramowski, and Kristian Kersting. 2019.
\newblock Pad{\'e} activation units: End-to-end learning of flexible activation functions in deep networks.
\newblock \emph{ArXiv}, abs/1907.06732.

\bibitem[{Moosavi et~al.(2022)Moosavi, Delfosse, Kersting, and Gurevych}]{Moosavi2022AdaptableA}
Nafise~Sadat Moosavi, Quentin Delfosse, Kristian Kersting, and Iryna Gurevych. 2022.
\newblock Adaptable adapters.
\newblock In \emph{North American Chapter of the Association for Computational Linguistics}.

\bibitem[{Novikova et~al.(2017)Novikova, Du{\v{s}}ek, and Rieser}]{novikova-etal-2017-e2e}
Jekaterina Novikova, Ond{\v{r}}ej Du{\v{s}}ek, and Verena Rieser. 2017.
\newblock \href {https://doi.org/10.18653/v1/W17-5525} {The {E}2{E} dataset: New challenges for end-to-end generation}.
\newblock In \emph{Proceedings of the 18th Annual {SIG}dial Meeting on Discourse and Dialogue}, pages 201--206, Saarbr{\"u}cken, Germany. Association for Computational Linguistics.

\bibitem[{{OpenAI}(2023)}]{gpt4}
{OpenAI}. 2023.
\newblock \href {https://doi.org/10.48550/arXiv.2303.08774} {{GPT-4 Technical Report}}.
\newblock \emph{arXiv e-prints}, page arXiv:2303.08774.

\bibitem[{Ouyang et~al.(2022)Ouyang, Wu, Jiang, Almeida, Wainwright, Mishkin, Zhang, Agarwal, Slama, Ray et~al.}]{ouyang2022training}
Long Ouyang, Jeffrey Wu, Xu~Jiang, Diogo Almeida, Carroll Wainwright, Pamela Mishkin, Chong Zhang, Sandhini Agarwal, Katarina Slama, Alex Ray, et~al. 2022.
\newblock Training language models to follow instructions with human feedback.
\newblock \emph{Advances in Neural Information Processing Systems}, 35:27730--27744.

\bibitem[{Pfeiffer et~al.(2021)Pfeiffer, Kamath, R{\"u}ckl{\'e}, Cho, and Gurevych}]{pfeiffer-etal-2021-adapterfusion}
Jonas Pfeiffer, Aishwarya Kamath, Andreas R{\"u}ckl{\'e}, Kyunghyun Cho, and Iryna Gurevych. 2021.
\newblock \href {https://doi.org/10.18653/v1/2021.eacl-main.39} {{A}dapter{F}usion: Non-destructive task composition for transfer learning}.
\newblock In \emph{Proceedings of the 16th Conference of the European Chapter of the Association for Computational Linguistics: Main Volume}, pages 487--503, Online. Association for Computational Linguistics.

\bibitem[{Qin et~al.(2023)Qin, Zhang, Zhang, Chen, Yasunaga, and Yang}]{qin2023chatgpt}
Chengwei Qin, Aston Zhang, Zhuosheng Zhang, Jiaao Chen, Michihiro Yasunaga, and Diyi Yang. 2023.
\newblock Is chatgpt a general-purpose natural language processing task solver?
\newblock \emph{arXiv preprint arXiv:2302.06476}.

\bibitem[{Radford et~al.(2019)Radford, Wu, Child, Luan, Amodei, Sutskever et~al.}]{radford2019language}
Alec Radford, Jeffrey Wu, Rewon Child, David Luan, Dario Amodei, Ilya Sutskever, et~al. 2019.
\newblock Language models are unsupervised multitask learners.
\newblock \emph{OpenAI blog}, 1(8):9.

\bibitem[{Rajpurkar et~al.(2016)Rajpurkar, Zhang, Lopyrev, and Liang}]{rajpurkar-etal-2016-squad}
Pranav Rajpurkar, Jian Zhang, Konstantin Lopyrev, and Percy Liang. 2016.
\newblock \href {https://doi.org/10.18653/v1/D16-1264} {{SQ}u{AD}: 100,000+ questions for machine comprehension of text}.
\newblock In \emph{Proceedings of the 2016 Conference on Empirical Methods in Natural Language Processing}, pages 2383--2392, Austin, Texas. Association for Computational Linguistics.

\bibitem[{R{\"u}ckl{\'e} et~al.(2020)R{\"u}ckl{\'e}, Geigle, Glockner, Beck, Pfeiffer, Reimers, and Gurevych}]{Rckl2020AdapterDropOT}
Andreas R{\"u}ckl{\'e}, Gregor Geigle, Max Glockner, Tilman Beck, Jonas Pfeiffer, Nils Reimers, and Iryna Gurevych. 2020.
\newblock Adapterdrop: On the efficiency of adapters in transformers.
\newblock In \emph{Conference on Empirical Methods in Natural Language Processing}.

\bibitem[{Sanh et~al.(2021)Sanh, Webson, Raffel, Bach, Sutawika, Alyafeai, Chaffin, Stiegler, Scao, Raja, Dey, Bari, Xu, Thakker, Sharma, Szczechla, Kim, Chhablani, Nayak, Datta, Chang, Jiang, Wang, Manica, Shen, Yong, Pandey, Bawden, Wang, Neeraj, Rozen, Sharma, Santilli, F{\'e}vry, Fries, Teehan, Biderman, Gao, Bers, Wolf, and Rush}]{Sanh2021MultitaskPT}
Victor Sanh, Albert Webson, Colin Raffel, Stephen~H. Bach, Lintang Sutawika, Zaid Alyafeai, Antoine Chaffin, Arnaud Stiegler, Teven~Le Scao, Arun Raja, Manan Dey, M~Saiful Bari, Canwen Xu, Urmish Thakker, Shanya~Sharma Sharma, Eliza Szczechla, Taewoon Kim, Gunjan Chhablani, Nihal~V. Nayak, Debajyoti Datta, Jonathan~D. Chang, Mike Tian-Jian Jiang, Han Wang, Matteo Manica, Sheng Shen, Zheng-Xin Yong, Harshit Pandey, Rachel Bawden, Thomas Wang, Trishala Neeraj, Jos Rozen, Abheesht Sharma, Andrea Santilli, Thibault F{\'e}vry, Jason~Alan Fries, Ryan Teehan, Stella Biderman, Leo Gao, Tali Bers, Thomas Wolf, and Alexander~M. Rush. 2021.
\newblock \href {https://api.semanticscholar.org/CorpusID:239009562} {Multitask prompted training enables zero-shot task generalization}.
\newblock \emph{ArXiv}, abs/2110.08207.

\bibitem[{Sun et~al.(2020{\natexlab{a}})Sun, Xiao, Zhu, He, Zhang, Xu, Hou, Li, Ni, and Xie}]{Sun2020MedicalKG}
Haixia Sun, Jin Xiao, Wei Zhu, Yilong He, Sheng Zhang, Xiaowei Xu, Li~Hou, Jiao Li, Yuan Ni, and Guo~Tong Xie. 2020{\natexlab{a}}.
\newblock \href {https://api.semanticscholar.org/CorpusID:220747092} {Medical knowledge graph to enhance fraud, waste, and abuse detection on claim data: Model development and performance evaluation}.
\newblock \emph{JMIR Medical Informatics}, 8.

\bibitem[{Sun et~al.(2020{\natexlab{b}})Sun, Xiao, Zhu, He, Zhang, Xu, Hou, Li, Ni, and Xie}]{info:doi/10.2196/17653}
Haixia Sun, Jin Xiao, Wei Zhu, Yilong He, Sheng Zhang, Xiaowei Xu, Li~Hou, Jiao Li, Yuan Ni, and Guotong Xie. 2020{\natexlab{b}}.
\newblock \href {https://doi.org/10.2196/17653} {Medical knowledge graph to enhance fraud, waste, and abuse detection on claim data: Model development and performance evaluation}.
\newblock \emph{JMIR Med Inform}, 8(7):e17653.

\bibitem[{Sun et~al.(2022)Sun, Liu, Zhu, Geng, Wu, He, Ni, Xie, Huang, and Qiu}]{sun-etal-2022-simple}
Tianxiang Sun, Xiangyang Liu, Wei Zhu, Zhichao Geng, Lingling Wu, Yilong He, Yuan Ni, Guotong Xie, Xuanjing Huang, and Xipeng Qiu. 2022.
\newblock \href {https://doi.org/10.18653/v1/2022.findings-acl.189} {A simple hash-based early exiting approach for language understanding and generation}.
\newblock In \emph{Findings of the Association for Computational Linguistics: ACL 2022}, pages 2409--2421, Dublin, Ireland. Association for Computational Linguistics.

\bibitem[{Sung et~al.(2022)Sung, Cho, and Bansal}]{Sung2022LSTLS}
Yi-Lin Sung, Jaemin Cho, and Mohit Bansal. 2022.
\newblock Lst: Ladder side-tuning for parameter and memory efficient transfer learning.
\newblock \emph{ArXiv}, abs/2206.06522.

\bibitem[{Taori et~al.(2023)Taori, Gulrajani, Zhang, Dubois, Li, Guestrin, Liang, and Hashimoto}]{alpaca}
Rohan Taori, Ishaan Gulrajani, Tianyi Zhang, Yann Dubois, Xuechen Li, Carlos Guestrin, Percy Liang, and Tatsunori~B. Hashimoto. 2023.
\newblock Stanford alpaca: An instruction-following llama model.
\newblock \url{https://github.com/tatsu-lab/stanford_alpaca}.

\bibitem[{Touvron et~al.(2023)Touvron, Martin, Stone, Albert, Almahairi, Babaei, Bashlykov, Batra, Bhargava, Bhosale, Bikel, Blecher, Ferrer, Chen, Cucurull, Esiobu, Fernandes, Fu, Fu, Fuller, Gao, Goswami, Goyal, Hartshorn, Hosseini, Hou, Inan, Kardas, Kerkez, Khabsa, Kloumann, Korenev, Koura, Lachaux, Lavril, Lee, Liskovich, Lu, Mao, Martinet, Mihaylov, Mishra, Molybog, Nie, Poulton, Reizenstein, Rungta, Saladi, Schelten, Silva, Smith, Subramanian, Tan, Tang, Taylor, Williams, Kuan, Xu, Yan, Zarov, Zhang, Fan, Kambadur, Narang, Rodriguez, Stojnic, Edunov, and Scialom}]{Touvron2023Llama2O}
Hugo Touvron, Louis Martin, Kevin~R. Stone, Peter Albert, Amjad Almahairi, Yasmine Babaei, Nikolay Bashlykov, Soumya Batra, Prajjwal Bhargava, Shruti Bhosale, Daniel~M. Bikel, Lukas Blecher, Cristian~Cant{\'o}n Ferrer, Moya Chen, Guillem Cucurull, David Esiobu, Jude Fernandes, Jeremy Fu, Wenyin Fu, Brian Fuller, Cynthia Gao, Vedanuj Goswami, Naman Goyal, Anthony~S. Hartshorn, Saghar Hosseini, Rui Hou, Hakan Inan, Marcin Kardas, Viktor Kerkez, Madian Khabsa, Isabel~M. Kloumann, A.~V. Korenev, Punit~Singh Koura, Marie-Anne Lachaux, Thibaut Lavril, Jenya Lee, Diana Liskovich, Yinghai Lu, Yuning Mao, Xavier Martinet, Todor Mihaylov, Pushkar Mishra, Igor Molybog, Yixin Nie, Andrew Poulton, Jeremy Reizenstein, Rashi Rungta, Kalyan Saladi, Alan Schelten, Ruan Silva, Eric~Michael Smith, R.~Subramanian, Xia Tan, Binh Tang, Ross Taylor, Adina Williams, Jian~Xiang Kuan, Puxin Xu, Zhengxu Yan, Iliyan Zarov, Yuchen Zhang, Angela Fan, Melanie Kambadur, Sharan Narang, Aurelien Rodriguez, Robert Stojnic, Sergey Edunov, and
  Thomas Scialom. 2023.
\newblock \href {https://api.semanticscholar.org/CorpusID:259950998} {Llama 2: Open foundation and fine-tuned chat models}.
\newblock \emph{ArXiv}, abs/2307.09288.

\bibitem[{Vaswani et~al.(2017)Vaswani, Shazeer, Parmar, Uszkoreit, Jones, Gomez, Kaiser, and Polosukhin}]{Vaswani2017AttentionIA}
Ashish Vaswani, Noam~M. Shazeer, Niki Parmar, Jakob Uszkoreit, Llion Jones, Aidan~N. Gomez, Lukasz Kaiser, and Illia Polosukhin. 2017.
\newblock Attention is all you need.
\newblock \emph{ArXiv}, abs/1706.03762.

\bibitem[{Wang et~al.(2019)Wang, Pruksachatkun, Nangia, Singh, Michael, Hill, Levy, and Bowman}]{Wang2019SuperGLUEAS}
Alex Wang, Yada Pruksachatkun, Nikita Nangia, Amanpreet Singh, Julian Michael, Felix Hill, Omer Levy, and Samuel~R. Bowman. 2019.
\newblock Superglue: A stickier benchmark for general-purpose language understanding systems.
\newblock \emph{ArXiv}, abs/1905.00537.

\bibitem[{Wang et~al.(2018)Wang, Singh, Michael, Hill, Levy, and Bowman}]{Wang2018GLUEAM}
Alex Wang, Amanpreet Singh, Julian Michael, Felix Hill, Omer Levy, and Samuel~R. Bowman. 2018.
\newblock Glue: A multi-task benchmark and analysis platform for natural language understanding.
\newblock In \emph{BlackboxNLP@EMNLP}.

\bibitem[{Wang et~al.(2020)Wang, Zhu, Jiang, Zhang, Wang, Ni, Xie, and Xiao}]{Wang2020MiningIH}
Li~Wang, Wei Zhu, Sihang Jiang, Sheng Zhang, Keqiang Wang, Yuan Ni, Guo~Tong Xie, and Yanghua Xiao. 2020.
\newblock \href {https://api.semanticscholar.org/CorpusID:224281022} {Mining infrequent high-quality phrases from domain-specific corpora}.
\newblock \emph{Proceedings of the 29th ACM International Conference on Information \& Knowledge Management}.

\bibitem[{Wang et~al.(2023)Wang, Chen, Zhu, Ni, Xie, Yang, and Xiao}]{Wang2023MultitaskEL}
Xuwu Wang, Lihan Chen, Wei Zhu, Yuan Ni, Guo~Tong Xie, Deqing Yang, and Yanghua Xiao. 2023.
\newblock \href {https://api.semanticscholar.org/CorpusID:258975891} {Multi-task entity linking with supervision from a taxonomy}.
\newblock \emph{Knowledge and Information Systems}, 65:4335 -- 4358.

\bibitem[{Wei et~al.(2021)Wei, Bosma, Zhao, Guu, Yu, Lester, Du, Dai, and Le}]{Wei2021FinetunedLM}
Jason Wei, Maarten Bosma, Vincent Zhao, Kelvin Guu, Adams~Wei Yu, Brian Lester, Nan Du, Andrew~M. Dai, and Quoc~V. Le. 2021.
\newblock \href {https://api.semanticscholar.org/CorpusID:237416585} {Finetuned language models are zero-shot learners}.
\newblock \emph{ArXiv}, abs/2109.01652.

\bibitem[{Wolf et~al.(2020{\natexlab{a}})Wolf, Debut, Sanh, Chaumond, Delangue, Moi, Cistac, Rault, Louf, Funtowicz et~al.}]{wolf2020transformers}
Thomas Wolf, Lysandre Debut, Victor Sanh, Julien Chaumond, Clement Delangue, Anthony Moi, Pierric Cistac, Tim Rault, R{\'e}mi Louf, Morgan Funtowicz, et~al. 2020{\natexlab{a}}.
\newblock Transformers: State-of-the-art natural language processing.
\newblock In \emph{Proceedings of the 2020 conference on empirical methods in natural language processing: system demonstrations}, pages 38--45.

\bibitem[{Wolf et~al.(2020{\natexlab{b}})Wolf, Debut, Sanh, Chaumond, Delangue, Moi, Cistac, Rault, Louf, Funtowicz, Davison, Shleifer, von Platen, Ma, Jernite, Plu, Xu, Scao, Gugger, Drame, Lhoest, and Rush}]{wolf-etal-2020-transformers}
Thomas Wolf, Lysandre Debut, Victor Sanh, Julien Chaumond, Clement Delangue, Anthony Moi, Pierric Cistac, Tim Rault, Rémi Louf, Morgan Funtowicz, Joe Davison, Sam Shleifer, Patrick von Platen, Clara Ma, Yacine Jernite, Julien Plu, Canwen Xu, Teven~Le Scao, Sylvain Gugger, Mariama Drame, Quentin Lhoest, and Alexander~M. Rush. 2020{\natexlab{b}}.
\newblock \href {https://www.aclweb.org/anthology/2020.emnlp-demos.6} {Transformers: State-of-the-art natural language processing}.
\newblock In \emph{Proceedings of the 2020 Conference on Empirical Methods in Natural Language Processing: System Demonstrations}, pages 38--45, Online. Association for Computational Linguistics.

\bibitem[{Wu et~al.(2022)Wu, Wang, Gu, Hou, Dong, Vydiswaran, and Ma}]{Wu2022IDPGAI}
Zhuofeng Wu, Sinong Wang, Jiatao Gu, Rui Hou, Yuxiao Dong, V.~G.~Vinod Vydiswaran, and Hao Ma. 2022.
\newblock Idpg: An instance-dependent prompt generation method.
\newblock In \emph{North American Chapter of the Association for Computational Linguistics}.

\bibitem[{Xin et~al.(2024)Xin, Luo, Zhou, Du, Liu, Fan, Li, and Du}]{Xin2024ParameterEfficientFF}
Yi~Xin, Siqi Luo, Haodi Zhou, Junlong Du, Xiaohong Liu, Yue Fan, Qing Li, and Yuntao Du. 2024.
\newblock Parameter-efficient fine-tuning for pre-trained vision models: A survey.
\newblock \emph{ArXiv}, abs/2402.02242.

\bibitem[{Xu et~al.(2023)Xu, Xie, Qin, Tao, and Wang}]{Xu2023ParameterEfficientFM}
Lingling Xu, Haoran Xie, Si-Zhao~Joe Qin, Xiaohui Tao, and Fu~Lee Wang. 2023.
\newblock Parameter-efficient fine-tuning methods for pretrained language models: A critical review and assessment.
\newblock \emph{ArXiv}, abs/2312.12148.

\bibitem[{Yin et~al.(2024)Yin, Wang, Yang, Tan, Li, Zhu, and Sung}]{yin2024machine}
Huiming Yin, Kun Wang, Ruyu Yang, Yanfang Tan, Qiang Li, Wei Zhu, and Suzi Sung. 2024.
\newblock A machine learning model for predicting acute exacerbation of in-home chronic obstructive pulmonary disease patients.
\newblock \emph{Computer Methods and Programs in Biomedicine}, page 108005.

\bibitem[{Yue et~al.(2024)Yue, Wang, Zhu, Guan, Zheng, Wang, Sun, and Ma}]{yue2024tcmbench}
Wenjing Yue, Xiaoling Wang, Wei Zhu, Ming Guan, Huanran Zheng, Pengfei Wang, Changzhi Sun, and Xin Ma. 2024.
\newblock Tcmbench: A comprehensive benchmark for evaluating large language models in traditional chinese medicine.
\newblock \emph{arXiv preprint arXiv:2406.01126}.

\bibitem[{Zhang et~al.(2023{\natexlab{a}})Zhang, Tan, Dai, and Zhu}]{Zhang2023LECOIE}
Jingfang Zhang, Ming Tan, Pengyu Dai, and Wei-Guo Zhu. 2023{\natexlab{a}}.
\newblock \href {https://api.semanticscholar.org/CorpusID:259370796} {Leco: Improving early exiting via learned exits and comparison-based exiting mechanism}.
\newblock In \emph{Annual Meeting of the Association for Computational Linguistics}.

\bibitem[{Zhang et~al.(2023{\natexlab{b}})Zhang, Zhang, Shi, Chu, and Li}]{Zhang2023LoRAFAML}
Longteng Zhang, Lin Zhang, Shaohuai Shi, Xiaowen Chu, and Bo~Li. 2023{\natexlab{b}}.
\newblock \href {https://api.semanticscholar.org/CorpusID:260683267} {Lora-fa: Memory-efficient low-rank adaptation for large language models fine-tuning}.
\newblock \emph{ArXiv}, abs/2308.03303.

\bibitem[{Zhang et~al.(2023{\natexlab{c}})Zhang, Chen, Bukharin, He, Cheng, Chen, and Zhao}]{Zhang2023AdaptiveBA}
Qingru Zhang, Minshuo Chen, Alexander~W. Bukharin, Pengcheng He, Yu~Cheng, Weizhu Chen, and Tuo Zhao. 2023{\natexlab{c}}.
\newblock \href {https://api.semanticscholar.org/CorpusID:257631760} {Adaptive budget allocation for parameter-efficient fine-tuning}.
\newblock \emph{ArXiv}, abs/2303.10512.

\bibitem[{Zhang et~al.(2020)Zhang, Wu, Katiyar, Weinberger, and Artzi}]{Zhang2020RevisitingFB}
Tianyi Zhang, Felix Wu, Arzoo Katiyar, Kilian~Q. Weinberger, and Yoav Artzi. 2020.
\newblock Revisiting few-sample bert fine-tuning.
\newblock \emph{ArXiv}, abs/2006.05987.

\bibitem[{Zhang et~al.(2023{\natexlab{d}})Zhang, Tan, Zhang, and Zhu}]{Zhang2023NAGNERAU}
Xinpeng Zhang, Ming Tan, Jingfan Zhang, and Wei Zhu. 2023{\natexlab{d}}.
\newblock \href {https://api.semanticscholar.org/CorpusID:259370837} {Nag-ner: a unified non-autoregressive generation framework for various ner tasks}.
\newblock In \emph{Annual Meeting of the Association for Computational Linguistics}.

\bibitem[{Zhang et~al.(2023{\natexlab{e}})Zhang, Gao, Zhu, and Wang}]{Zhang2023FastNERSU}
Yuming Zhang, Xiangxiang Gao, Wei Zhu, and Xiaoling Wang. 2023{\natexlab{e}}.
\newblock \href {https://api.semanticscholar.org/CorpusID:265214231} {Fastner: Speeding up inferences for named entity recognition tasks}.
\newblock In \emph{International Conference on Advanced Data Mining and Applications}.

\bibitem[{Zhang et~al.(2023{\natexlab{f}})Zhang, Wang, Tan, and Zhu}]{Zhang2023LearnedAA}
Yuming Zhang, Peng Wang, Ming Tan, and Wei-Guo Zhu. 2023{\natexlab{f}}.
\newblock \href {https://api.semanticscholar.org/CorpusID:259858833} {Learned adapters are better than manually designed adapters}.
\newblock In \emph{Annual Meeting of the Association for Computational Linguistics}.

\bibitem[{Zhang et~al.(2022)Zhang, Zhu, Zhang, Wang, Jin, and Chung}]{zhang-etal-2022-pcee}
Zhen Zhang, Wei Zhu, Jinfan Zhang, Peng Wang, Rize Jin, and Tae-Sun Chung. 2022.
\newblock \href {https://doi.org/10.18653/v1/2022.findings-naacl.25} {{PCEE}-{BERT}: Accelerating {BERT} inference via patient and confident early exiting}.
\newblock In \emph{Findings of the Association for Computational Linguistics: NAACL 2022}, pages 327--338, Seattle, United States. Association for Computational Linguistics.

\bibitem[{Zhang et~al.(2021)Zhang, Zhu, Yan, Gao, and Xie}]{zhang2021automatic}
Zhexi Zhang, Wei Zhu, Junchi Yan, Peng Gao, and Guotong Xie. 2021.
\newblock Automatic student network search for knowledge distillation.
\newblock In \emph{2020 25th International Conference on Pattern Recognition (ICPR)}, pages 2446--2453. IEEE.

\bibitem[{Zhao et~al.(2020)Zhao, Lin, Mi, Jaggi, and Sch{\"u}tze}]{zhao-etal-2020-masking}
Mengjie Zhao, Tao Lin, Fei Mi, Martin Jaggi, and Hinrich Sch{\"u}tze. 2020.
\newblock \href {https://doi.org/10.18653/v1/2020.emnlp-main.174} {Masking as an efficient alternative to finetuning for pretrained language models}.
\newblock In \emph{Proceedings of the 2020 Conference on Empirical Methods in Natural Language Processing (EMNLP)}, pages 2226--2241, Online. Association for Computational Linguistics.

\bibitem[{{Zhao} et~al.(2023){Zhao}, {Zhou}, {Li}, {Tang}, {Wang}, {Hou}, {Min}, {Zhang}, {Zhang}, {Dong}, {Du}, {Yang}, {Chen}, {Chen}, {Jiang}, {Ren}, {Li}, {Tang}, {Liu}, {Liu}, {Nie}, and {Wen}}]{2023arXiv230318223Z}
Wayne~Xin {Zhao}, Kun {Zhou}, Junyi {Li}, Tianyi {Tang}, Xiaolei {Wang}, Yupeng {Hou}, Yingqian {Min}, Beichen {Zhang}, Junjie {Zhang}, Zican {Dong}, Yifan {Du}, Chen {Yang}, Yushuo {Chen}, Zhipeng {Chen}, Jinhao {Jiang}, Ruiyang {Ren}, Yifan {Li}, Xinyu {Tang}, Zikang {Liu}, Peiyu {Liu}, Jian-Yun {Nie}, and Ji-Rong {Wen}. 2023.
\newblock \href {https://doi.org/10.48550/arXiv.2303.18223} {{A Survey of Large Language Models}}.
\newblock \emph{arXiv e-prints}, page arXiv:2303.18223.

\bibitem[{Zheng et~al.(2023)Zheng, Zhu, Wang, and Wang}]{Zheng2023CandidateSF}
Huanran Zheng, Wei Zhu, Pengfei Wang, and Xiaoling Wang. 2023.
\newblock \href {https://api.semanticscholar.org/CorpusID:256358677} {Candidate soups: Fusing candidate results improves translation quality for non-autoregressive translation}.
\newblock \emph{ArXiv}, abs/2301.11503.

\bibitem[{Zheng et~al.(2024)Zheng, Zhu, and Wang}]{zheng2024nat4at}
Huanran Zheng, Wei Zhu, and Xiaoling Wang. 2024.
\newblock Nat4at: Using non-autoregressive translation makes autoregressive translation faster and better.
\newblock In \emph{Proceedings of the ACM on Web Conference 2024}, pages 4181--4192.

\bibitem[{{Zheng} et~al.(2023){Zheng}, {Chiang}, {Sheng}, {Zhuang}, {Wu}, {Zhuang}, {Lin}, {Li}, {Li}, {Xing}, {Zhang}, {Gonzalez}, and {Stoica}}]{2023arXiv230605685Z}
Lianmin {Zheng}, Wei-Lin {Chiang}, Ying {Sheng}, Siyuan {Zhuang}, Zhanghao {Wu}, Yonghao {Zhuang}, Zi~{Lin}, Zhuohan {Li}, Dacheng {Li}, Eric.~P {Xing}, Hao {Zhang}, Joseph~E. {Gonzalez}, and Ion {Stoica}. 2023.
\newblock \href {https://doi.org/10.48550/arXiv.2306.05685} {{Judging LLM-as-a-Judge with MT-Bench and Chatbot Arena}}.
\newblock \emph{arXiv e-prints}, page arXiv:2306.05685.

\bibitem[{Zhong et~al.(2017)Zhong, Xiong, and Socher}]{Zhong2017Seq2SQLGS}
Victor Zhong, Caiming Xiong, and Richard Socher. 2017.
\newblock Seq2sql: Generating structured queries from natural language using reinforcement learning.
\newblock \emph{ArXiv}, abs/1709.00103.

\bibitem[{Zhou et~al.(2019)Zhou, Ni, Xie, Zhu, Chen, Wang, and Pan}]{zhou2019analysis}
Xiaofeng Zhou, Yuan Ni, Guotong Xie, Wei Zhu, Cai Chen, Tianhao Wang, and Zhigang Pan. 2019.
\newblock Analysis of the health information needs of diabetics in china.
\newblock In \emph{MEDINFO 2019: Health and Wellbeing e-Networks for All}, pages 487--491. IOS Press.

\bibitem[{Zhu(2021{\natexlab{a}})}]{zhu2021autonlu}
Wei Zhu. 2021{\natexlab{a}}.
\newblock Autonlu: Architecture search for sentence and cross-sentence attention modeling with re-designed search space.
\newblock In \emph{Natural Language Processing and Chinese Computing: 10th CCF International Conference, NLPCC 2021, Qingdao, China, October 13--17, 2021, Proceedings, Part I 10}, pages 155--168. Springer.

\bibitem[{Zhu(2021{\natexlab{b}})}]{zhu-2021-autorc}
Wei Zhu. 2021{\natexlab{b}}.
\newblock \href {https://doi.org/10.18653/v1/2021.acl-srw.4} {{A}uto{RC}: Improving {BERT} based relation classification models via architecture search}.
\newblock In \emph{Proceedings of the 59th Annual Meeting of the Association for Computational Linguistics and the 11th International Joint Conference on Natural Language Processing: Student Research Workshop}, pages 33--43, Online. Association for Computational Linguistics.

\bibitem[{Zhu(2021{\natexlab{c}})}]{zhu-2021-leebert}
Wei Zhu. 2021{\natexlab{c}}.
\newblock \href {https://doi.org/10.18653/v1/2021.acl-long.231} {{L}ee{BERT}: Learned early exit for {BERT} with cross-level optimization}.
\newblock In \emph{Proceedings of the 59th Annual Meeting of the Association for Computational Linguistics and the 11th International Joint Conference on Natural Language Processing (Volume 1: Long Papers)}, pages 2968--2980, Online. Association for Computational Linguistics.

\bibitem[{Zhu(2021{\natexlab{d}})}]{Zhu2021MVPBERTMP}
Wei Zhu. 2021{\natexlab{d}}.
\newblock \href {https://api.semanticscholar.org/CorpusID:237331564} {Mvp-bert: Multi-vocab pre-training for chinese bert}.
\newblock In \emph{Annual Meeting of the Association for Computational Linguistics}.

\bibitem[{Zhu et~al.(2021{\natexlab{a}})Zhu, He, Chai, Fan, Ni, Xie, and Wang}]{zhu_etal_2021_paht}
Wei Zhu, Yilong He, Ling Chai, Yuanchun Fan, Yuan Ni, Guo~Tong Xie, and Xiaoling Wang. 2021{\natexlab{a}}.
\newblock \href {https://api.semanticscholar.org/CorpusID:235097590} {paht\_nlp @ mediqa 2021: Multi-grained query focused multi-answer summarization}.
\newblock In \emph{Workshop on Biomedical Natural Language Processing}.

\bibitem[{Zhu et~al.(2024)Zhu, Li, Tian, Wang, Wang, Chen, Wu, Ni, and Xie}]{zhu2024text2mdt}
Wei Zhu, Wenfeng Li, Xing Tian, Pengfei Wang, Xiaoling Wang, Jin Chen, Yuanbin Wu, Yuan Ni, and Guotong Xie. 2024.
\newblock Text2mdt: Extracting medical decision trees from medical texts.
\newblock \emph{arXiv preprint arXiv:2401.02034}.

\bibitem[{Zhu et~al.(2023{\natexlab{a}})Zhu, Li, Wang, Ji, Wu, Chen, Chen, and Tang}]{text2dt_shared_task}
Wei Zhu, Wenfeng Li, Xiaoling Wang, Wendi Ji, Yuanbin Wu, Jin Chen, Liang Chen, and Buzhou Tang. 2023{\natexlab{a}}.
\newblock Extracting decision trees from medical texts: An overview of the text2dt track in chip2022.
\newblock In \emph{Health Information Processing. Evaluation Track Papers}, pages 89--102, Singapore. Springer Nature Singapore.

\bibitem[{Zhu et~al.(2023{\natexlab{b}})Zhu, Li, Wang, Ji, Wu, Chen, Chen, and Tang}]{Text2dt}
Wei Zhu, Wenfeng Li, Xiaoling Wang, Wendi Ji, Yuanbin Wu, Jin Chen, Liang Chen, and Buzhou Tang. 2023{\natexlab{b}}.
\newblock Extracting decision trees from medical texts: An overview of the text2dt track in chip2022.
\newblock In \emph{Health Information Processing. Evaluation Track Papers}, pages 89--102, Singapore. Springer Nature Singapore.

\bibitem[{Zhu et~al.(2021{\natexlab{b}})Zhu, Ni, Wang, and Xie}]{zhu-etal-2021-discovering}
Wei Zhu, Yuan Ni, Xiaoling Wang, and Guotong Xie. 2021{\natexlab{b}}.
\newblock \href {https://doi.org/10.18653/v1/2021.naacl-industry.29} {Discovering better model architectures for medical query understanding}.
\newblock In \emph{Proceedings of the 2021 Conference of the North American Chapter of the Association for Computational Linguistics: Human Language Technologies: Industry Papers}, pages 230--237, Online. Association for Computational Linguistics.

\bibitem[{Zhu et~al.(2019{\natexlab{a}})Zhu, Ni, Xie, Zhou, and Chen}]{Zhu2019TheDS}
Wei Zhu, Yuan Ni, Guo~Tong Xie, Xiaofeng Zhou, and Cai Chen. 2019{\natexlab{a}}.
\newblock \href {https://api.semanticscholar.org/CorpusID:208207213} {The dr-kgqa system for automatically answering medication related questions in chinese}.
\newblock \emph{2019 IEEE International Conference on Healthcare Informatics (ICHI)}, pages 1--6.

\bibitem[{Zhu and Tan(2023{\natexlab{a}})}]{zhu2023spt}
Wei Zhu and Ming Tan. 2023{\natexlab{a}}.
\newblock Spt: Learning to selectively insert prompts for better prompt tuning.
\newblock In \emph{The 2023 Conference on Empirical Methods in Natural Language Processing}.

\bibitem[{Zhu and Tan(2023{\natexlab{b}})}]{zhu-tan-2023-spt}
Wei Zhu and Ming Tan. 2023{\natexlab{b}}.
\newblock \href {https://aclanthology.org/2023.emnlp-main.727} {{SPT}: Learning to selectively insert prompts for better prompt tuning}.
\newblock In \emph{Proceedings of the 2023 Conference on Empirical Methods in Natural Language Processing}, pages 11862--11878, Singapore. Association for Computational Linguistics.

\bibitem[{Zhu et~al.(2023{\natexlab{c}})Zhu, Wang, Ni, Xie, and Wang}]{Zhu2023BADGESU}
Wei Zhu, Peifeng Wang, Yuan Ni, Guo~Tong Xie, and Xiaoling Wang. 2023{\natexlab{c}}.
\newblock \href {https://api.semanticscholar.org/CorpusID:259370582} {Badge: Speeding up bert inference after deployment via block-wise bypasses and divergence-based early exiting}.
\newblock In \emph{Annual Meeting of the Association for Computational Linguistics}.

\bibitem[{Zhu et~al.(2023{\natexlab{d}})Zhu, Wang, Wang, Ni, and Xie}]{ACF}
Wei Zhu, Peng Wang, Xiaoling Wang, Yuan Ni, and Guotong Xie. 2023{\natexlab{d}}.
\newblock \href {https://doi.org/10.1109/ICASSP49357.2023.10094859} {Acf: Aligned contrastive finetuning for language and vision tasks}.
\newblock In \emph{ICASSP 2023 - 2023 IEEE International Conference on Acoustics, Speech and Signal Processing (ICASSP)}, pages 1--5.

\bibitem[{Zhu et~al.(2023{\natexlab{e}})Zhu, Wang, Chen, and Tang}]{Zhu2023OverviewOT}
Wei Zhu, Xiaoling Wang, Mosha Chen, and Buzhou Tang. 2023{\natexlab{e}}.
\newblock \href {https://api.semanticscholar.org/CorpusID:266690968} {Overview of the promptcblue shared task in chip2023}.
\newblock \emph{ArXiv}, abs/2312.17522.

\bibitem[{Zhu et~al.(2023{\natexlab{f}})Zhu, Wang, Chen, and Tang}]{zhu2023overview}
Wei Zhu, Xiaoling Wang, Mosha Chen, and Buzhou Tang. 2023{\natexlab{f}}.
\newblock Overview of the promptcblue shared task in chip2023.
\newblock In \emph{China Health Information Processing Conference}, pages 3--20. Springer.

\bibitem[{Zhu et~al.(2021{\natexlab{c}})Zhu, Wang, Ni, and Xie}]{autotrans}
Wei Zhu, Xiaoling Wang, Yuan Ni, and Guotong Xie. 2021{\natexlab{c}}.
\newblock Autotrans: Automating transformer design via reinforced architecture search.
\newblock In \emph{Natural Language Processing and Chinese Computing}, pages 169--182, Cham. Springer International Publishing.

\bibitem[{Zhu et~al.(2021{\natexlab{d}})Zhu, Wang, Ni, and Xie}]{zhu-etal-2021-gaml}
Wei Zhu, Xiaoling Wang, Yuan Ni, and Guotong Xie. 2021{\natexlab{d}}.
\newblock \href {https://aclanthology.org/2021.emnlp-main.242} {{GAML}-{BERT}: Improving {BERT} early exiting by gradient aligned mutual learning}.
\newblock In \emph{Proceedings of the 2021 Conference on Empirical Methods in Natural Language Processing}, pages 3033--3044, Online and Punta Cana, Dominican Republic. Association for Computational Linguistics.

\bibitem[{{Zhu} et~al.(2023){Zhu}, {Wang}, {Zheng}, {Chen}, and {Tang}}]{PromptCBLUE}
Wei {Zhu}, Xiaoling {Wang}, Huanran {Zheng}, Mosha {Chen}, and Buzhou {Tang}. 2023.
\newblock \href {https://doi.org/10.48550/arXiv.2310.14151} {{PromptCBLUE: A Chinese Prompt Tuning Benchmark for the Medical Domain}}.
\newblock \emph{arXiv e-prints}, page arXiv:2310.14151.

\bibitem[{Zhu et~al.(2019{\natexlab{b}})Zhu, Zhou, Wang, Luo, Li, Ni, and Xie}]{zhu-etal-2019-panlp}
Wei Zhu, Xiaofeng Zhou, Keqiang Wang, Xun Luo, Xiepeng Li, Yuan Ni, and Guotong Xie. 2019{\natexlab{b}}.
\newblock \href {https://doi.org/10.18653/v1/W19-5040} {{PANLP} at {MEDIQA} 2019: Pre-trained language models, transfer learning and knowledge distillation}.
\newblock In \emph{Proceedings of the 18th BioNLP Workshop and Shared Task}, pages 380--388, Florence, Italy. Association for Computational Linguistics.

\bibitem[{Zuo et~al.(2022)Zuo, Zhu, and Cai}]{zuo-etal-2022-continually}
Yuhui Zuo, Wei Zhu, and Guoyong~GUET Cai. 2022.
\newblock \href {https://aclanthology.org/2022.coling-1.268} {Continually detection, rapidly react: Unseen rumors detection based on continual prompt-tuning}.
\newblock In \emph{Proceedings of the 29th International Conference on Computational Linguistics}, pages 3029--3041, Gyeongju, Republic of Korea. International Committee on Computational Linguistics.

\end{thebibliography}
\bibliographystyle{acl_natbib}

\appendix

\section{Additional related works}

\subsection{Adapter-based tuning.} 

One of the most important research lines of PEFT is adapter-based tuning. Adapter \cite{houlsby2019parameter} inserts adapter modules with bottleneck architecture between every consecutive Transformer \cite{Vaswani2017AttentionIA} sublayers. AdapterFusion \cite{pfeiffer-etal-2021-adapterfusion} only inserts sequential adapters after the feed-forward module. Adapter-based tuning methods have comparable results with model tuning when only tuning a fraction of the backbone model's parameter number. Due to their strong performance, a branch of literature has investigated the architecture of adapters in search of further improvements. \citet{He2021TowardsAU} analyze a wide range of PETuning methods and show that they are essentially equivalent. They also propose the general architecture of PEFT, and derive the Parallel Adapter which connects the adapter modules in parallel to the self-attention and MLP modules in the Transformer block. AdapterDrop \cite{Rckl2020AdapterDropOT} investigates the efficiency of removing adapters from lower layers. Adaptive adapters \cite{Moosavi2022AdaptableA} investigate the activation functions of adapters and propose to learn the activation functions of adapters via optimizing the parameters of rational functions as a part of the model parameters. Compacter \cite{Mahabadi2021CompacterEL} uses low-rank parameterized hypercomplex multiplication \cite{Le2021ParameterizedHG} to compress adapters' tunable parameters. LST \cite{Sung2022LSTLS} improves the memory efficiency by forming the adapters as a ladder along stacked Transformer blocks, and it enhances the adapter module by adding a self-attention module to its bottleneck architecture. \cite{Sung2022LSTLS,Jie2022ConvolutionalBA} try to add different encoding operations, like self-attention operations and convolutions between the bottleneck structure of adapters, and achieve better performances. Learned-Adapter \cite{Zhang2023LearnedAA} builds upon the above adapter-based methods and enhance the performance of adapter tuning by automatically learning better architectures for adapters.

\subsection{Literature on the LoRA methods}  

Since LoRA is the most popular PEFT method in the era of large language models, there are many works that are orthogonal to AdaLoRA, SoRA and our work that are devoted to improve LoRA on many different aspects. QLoRA \cite{2023arXiv230514314D} proposes a novel quantization method that can significantly reduce the memory consumptions of LLMs during LoRA fine-tuning. LoRA-FA \cite{Zhang2023LoRAFAML} freezes parts of the randomly initialized LoRA matrices. (d) VERA \cite{Kopiczko2023VeRAVR} investigate whether one could froze the randomly initialized LoRA matrices and only learns a set of scaling vectors. Tying LoRA matrices across layers are also investigated by VERA.

\section{Appendix: introduction to bi-level optimization}
\label{sec:appendix_bi_level_opt}

The bi-level optimization \cite{Liu2019DARTSDA} optimize $\Theta$ conditioned on the optimized parameters of $\Omega^{*}$. Denote the training set as $\mathcal{D}_{train}$, and the validation set as $\mathcal{D}_{val}$. The inner and outer levels of optimization are conducted on these two separate splits of the task dataset, which is analogous to validating architectures trained on $\mathcal{D}_{train}$ using a different split $\mathcal{D}_{val}$ to avoid over-fitting. Thus the optimization objective is:
\begin{align}
& \min_{\Theta} \mathcal{L}(\mathcal{D}_{val}, \Omega^{*}, \Theta),  \nonumber\\
& \emph{s.t.} \ \ \Omega^{*} =  \arg\min_{\Omega} \mathcal{L}(\mathcal{D}_{train}, \Omega, \Theta), 
\label{eq:bi_level_optimize}
\end{align}
where $\mathcal{L}()$ is the objective function on a given downstream task, such as cross entropy loss. The above bi-level optimization problem is approximated with an alternating optimization strategy. The gradients of $\Omega$ are calculated with batches of samples from $\mathcal{D}_{train}$, and the gradients of $\Theta$ are calculated on $\mathcal{D}_{val}$.

\section{Appendix for the datsets and evaluation metrics}
\label{sec:appendix_datasets}

\subsection{Datasets from GLUE and SuperGLUE }

We experiment on three tasks from the GLUE \cite{Wang2018GLUEAM} benchmark: (a) (a) a sentiment classification task, SST-2. (b) two benchmark natural language inference tasks, RTE and QNLI. We also experiment with three question-answering tasks: (a) two question answering tasks in the format of binary choices, COPA and BoolQ. (b) A SQuAD \cite{rajpurkar-etal-2016-squad} style question answering task, ReCoRD.

Since the original test sets are not publicly available for these tasks, we follow \citet{Zhang2020RevisitingFB,Mahabadi2021CompacterEL} to construct the train/dev/test splits as follows to ensure a fiar comparison: (a) for datasets with fewer than 10k samples (RTE, COPA, BoolQ), we divide the original validation set in half, using one half for validation and the other for testing. (b) for larger datasets, we split 1k samples from the training set as the development set, and use the original development set as the test set. The detailed statistics of the GLUE and SuperGLUE benchmark tasks is presented in Table \ref{tab:dataset_stats}.

\begin{table*}[tb!]
\centering
\resizebox{1.0\textwidth}{!}{
\begin{tabular}{cccccccc}
\hline
Datasets  &  \#train    &  \#dev   &   \#test   &   $ | \mathcal{Y} | $   &   Type   &  Labels  &  Metrics  \\ 
\hline
\multicolumn{8}{c}{\textbf{\emph{SuperGLUE tasks}}}  \\
\hline
BoolQ  &  9.4k   &    1.6k   &  1.6k   &    2   &   Question Answering    &  True, False   &   acc \\
COPA   &   0.4k   &   0.05k   &   0.05k    &   2   &      Question Answering   &    choice1, choice2   &   acc   \\
ReCoRD  &    101k &    1k   &    7.4k  &  - 
 &  Question Answering   &  -   &  f1-em  \\
\hline
\multicolumn{8}{c}{\textbf{\emph{GLUE tasks}}}  \\
\hline
SST-2  &  66k  &   1k    &   0.8k   &   2   &  sentiment classification &  positive, negative    &  acc     \\
RTE &   2.5k   &   0.1k   &    0.1k  &    2   &    NLI   &    entailment, not entailment  &  acc    \\
QNLI &  104k   &    1k   &    5.4k  &   2   &   NLI   &   entailment, not entailment   &  acc  \\

\hline
\multicolumn{8}{c}{\textbf{\emph{Other tasks}}}  \\
\hline

SQuAD &   87k  &  1k    &   5.9k   &   -   &  Question Answering 
 &  -   &   f1-em   \\
E2E &   42k  &  4.6k   &  4.6k   &   -   &  NLG   &  -    &  rouge-l   \\

GSM8K  &   7K  &  0.5K  &  1K   &  - &   Math reasoning  &   -    &   acc  \\

WikiSQL  &   61k  &  9K  &  17K   &   -  &   SQL generation  &   -    &   acc  \\

Alpaca   &    50k  &    1k    &  -   &  -  &  Instruction tuning  &  -   &   -   \\
MT-Bench   &   -   &  -  &   80  &   -   &    Instruction tuning  &  -    &  GPT-4 scores       \\

\hline
\end{tabular}}
\caption{\label{tab:dataset_stats}  The dataset statistics of the GLUE and SuperGLUE benchmark tasks evaluated in this work. $ | \mathcal{Y} | $ is the number of classes for a classification task. }
\end{table*}

\subsection{The SQuAD task}

Stanford Question Answering Dataset (SQuAD) \cite{rajpurkar-etal-2016-squad} is a reading comprehension dataset, consisting of questions posed by crowdworkers on a set of Wikipedia articles, where the answer to every question is a segment of text, or span, from the corresponding reading passage, or the question might be unanswerable. This task is one of the most widely studied question answering task in the field. 

In this work, we use the v1.1 version of SQuAD. Since the original test sets are not publicly available for these tasks, we follow \citet{Zhang2020RevisitingFB,Mahabadi2021CompacterEL} and split 1k samples from the training set as the development set, and use the original development set as the test set. The detailed statistics of this task is presented in Table \ref{tab:dataset_stats}.

\subsection{E2E benchmark}

The E2E benchmark dataset for training end-to-end, data-driven natural language generation systems in the restaurant domain. It asks a model to generate natural utterances based on a set of given key contents. This dataset has a 42061/4672/4693 train/dev/test split.

\subsection{GSM8K benchmark}

GSM8K is a dataset of 8.5K high quality linguistically diverse grade school math word problems created by human problem writers. The dataset is segmented into 7.5K training problems and 1K test problems. These problems take between 2 and 8 steps to solve, and solutions primarily involve performing a sequence of elementary calculations using basic arithmetic operations ($+-\times\div$) to reach the final answer. A bright middle school student should be able to solve every problem. It can be used for multi-step mathematical reasoning. We randomly select 0.5k samples from the training set to be the dev set.

\subsection{WikiSQL dataset}

WikiSQL consists of a corpus of 87,726 hand-annotated SQL query and natural language question pairs. These SQL queries are further split into training (61,297 examples), development (9,145 examples) and test sets (17,284 examples). It can be used for natural language inference tasks related to relational databases. In this work, we will ask the LLMs to generate SQL queries based on the given natural language questions.

\subsection{Instruction tuning}

Instruction tuning is an important method to improve the general capabilities of large language models \cite{ouyang2022training}. With the rise of large language models in the scale of 10B parameters or more,  like GPT-3, T5, PaLM, researchers have actively explored the few-shot or zero-shot capabilities of these models. \cite{Mishra2021CrossTaskGV} find that fine-tuning these LLMs on a large scale datasets containing hundreds of NLP tasks significantly improves the zero-shot performances on unseen tasks, establishing the scaling law of task numbers. The previous works like \cite{Wei2021FinetunedLM} and T0 \cite{Sanh2021MultitaskPT} establishes the instruction tuning datasets by transforming the traditional NLP tasks into a unified prompt format. Instruct-GPT \cite{ouyang2022training} conducts instruction tuning using the dataset constructed based the user queries from the OpenAI API users. Note that this work is also a seminal work for human feedback learning with reinforcement learning. However, the complete instruction tuning dataset from \cite{ouyang2022training} remains closed. With the launch of ChatGPT, \cite{alpaca} (Alpaca) constructs an instruction tuning dataset with diverse topics using the self-instruct techniques. 

 For our experiment, we employ the Alpaca dataset \cite{alpaca} for instruction tuning. Specifically, we employs its cleaned version\footnote{\url{https://huggingface.co/datasets/yahma/alpaca-cleaned}.}. This dataset comprises 51K instructions and demonstrations, and is suitable for instruction tuning. The cleaned version corrects multiple issues such as hallucinations, merged instructions, and empty outputs. We set aside 1000 samples in the Alpaca dataset as the validation set.

\subsection{Evaluation metrics/protocols}
\label{sec:appendix_evaluations}

For the three GLUE tasks we experiment on, we report accuracy (denoted as acc). For ReCoRD, we report the average of the F1 score and the exact match score (denoted as f1-em). For the BoolQ and COPA tasks, we report accuracy. The above choices of evaluation metrics strictly follow \cite{Wang2018GLUEAM} and \cite{Wang2019SuperGLUEAS}. 

For the SQuAD dataset, we also report the average of the F1 score and the exact match score (denoted as f1-em).

For the E2E benchmark, following \cite{novikova-etal-2017-e2e}, we report the ROUGE-L metric (denoted as rouge-l). We rely on the HuggingFace Evaluate package\footnote{\url{https://huggingface.co/docs/evaluate/index}} for computing this metric.

For the GSM8K task, we will directly consider the correctness of the final answers. Thus, we report accuracy (denoted as acc). 

For the WikiSQL, we will consider the correctness of the generated SQL queries. A predicted SQL query is correct if and only if it can be executed and obtains the same results with the ground truth.

For evaluating the quality of instruction tuned LlaMA-2 7B, we follow the current common practice of utilizing GPT-4 as a unbiased reviewer \cite{2023arXiv230605685Z}. 80 instructions from the MT-Bench is set as a test set. We generate model responses from a fine-tuned model with beam size 5 with the generation function in Huggingface Transformers \cite{wolf2020transformers}. Then we compare AdaLoRA and IAPT's answers with GPT-4. For each instruction in MT-Bench, GPT-4 \cite{gpt4} is asked to write a review for both answers from the two methods, and assigns a quantitative score on a scale of 10 to each response. The prompts of instructing GPT-4 for evaluation is presented in Appendix \ref{sec:appendix_gpt4_eval}. ROUGE-L scores computed by considering the answers generated by GPT-4 as the ground truth are also reported.

\section{Prompt templates for fine-tuning LlaMA-2 7B}
\label{sec:appendix_prompt_templates}

Since we fine-tune LlaMA-2 7B without introducing task-specific prediction heads, we need to transform all the tasks into a prompt-response format. First, following LlaMA-2 \cite{Touvron2023Llama2O}, we use a system prompting template, in which <query> denotes the user input, <response> denotes the assistants' targeted responses. All the samples will be input into this template before being fed to the LLMs. 
\begin{verbatim} 
<s>[INST] <<SYS>>
You are a helpful, respectful and honest 
assistant.
<</SYS>>

<query>[/INST]<response></s>
\end{verbatim}

Now we present the prompt-response template for each task. 

\noindent \textbf{Templates for RTE and QNLI} Since these two tasks are NLI tasks, the samples in them consists of two input text, [sentence1] and [sentence1], and a label [label\_name] (entailment or not entailment). Thus, we use the following templates:

Template for prompt: 
\begin{verbatim} 
sentence 1: [sentence1]
sentence 2: [sentence1]
Are sentence 1 and sentence 2 have 
entailment relation or not?
\end{verbatim}

Template for response:
\begin{verbatim} 
[label_name]
\end{verbatim}

\noindent \textbf{Templates for SST-2} The samples in this task consists of one input text, [sentence], and a label [label\_name] (positive or negative). 

Template for prompt: 
\begin{verbatim} 
[sentence] 
The sentiment of the given sentence is:
\end{verbatim}

Template for response:
\begin{verbatim} 
[label_name]
\end{verbatim}

\noindent \textbf{Templates for BoolQ} The samples in this task consists of a reference document, [doc], a query, [query], and a label [label\_name] (yes or no). 

Template for prompt: 
\begin{verbatim} 
Reference document:
[doc]
Question:
[query]
\end{verbatim}

Template for response:
\begin{verbatim} 
[label_name]
\end{verbatim}

\noindent \textbf{Templates for COPA} The samples in this task consists of a premise, [premise], two choices, [choice1] and [choice2], a query, [query], and a label [label\_name] (1 or 2, indicating which choice is consistent with the premise). 

Template for prompt: 
\begin{verbatim} 
Premise:
[premise]
Choice 1: [choice1]
Choice 2: [choice2]
Question:
[query]
\end{verbatim}

Template for response:
\begin{verbatim} 
[label_name]
\end{verbatim}

\noindent \textbf{Templates for ReCoRD and SQuAD} The samples in these two tasks consist of a context document, [context], a question, [query], and a answering span, [answer]. 

Template for prompt: 
\begin{verbatim} 
Context:
[context]
Question:
[query]
\end{verbatim}

Template for response:
\begin{verbatim} 
[answer]
\end{verbatim}

\noindent \textbf{Templates for E2E} The samples in this task consists of a reference [ref], consisting required information, and a targeted response, [target], which is a customer review written according to the reference's contents. 

Template for prompt: 
\begin{verbatim} 
Reference:
[ref]
Generate a customer review following the 
given reference.
\end{verbatim}

Template for response:
\begin{verbatim} 
[target]
\end{verbatim}

\noindent \textbf{Templates for GSM8K} The samples in this task consists of a math question [question], and a targeted response, [target] which is the reasoning or calculation steps for the math question. 

Template for prompt: 
\begin{verbatim} 
Answer the following math quesition:
[ref]
Instruction: please think step by step.
\end{verbatim}

Template for response:
\begin{verbatim} 
[target]
\end{verbatim}

\noindent \textbf{Templates for WikiSQL} The samples in this task consists of a natural language query [query], and information for the SQL table [table\_info], and a targeted response containing the SQL query, [target] which is the reasoning or calculation steps for the math question. 

Template for prompt: 
\begin{verbatim} 
Answer the following query by writing a 
SQL query on the given table:
[query]
Table information: 
[table\_info].
\end{verbatim}

Template for response:
\begin{verbatim} 
[target]
\end{verbatim}

\section{Prompt templates for GPT-4 evaluations}
\label{sec:appendix_gpt4_eval}
In this work, we utilize the powerful LLM GPT-4 \cite{gpt4} as the evaluator for comparing the instruction tuning quality. As a reviewer, GPT-4 will receive a query [query], two responses, [response1] and [response2], from two assistants. We will ask GPT-4 to write a review for each response, assessing the quality of the response, and then ask GPT-4 to assign a score on a scale of 10 to each response.

Template for prompt: 
\begin{verbatim} 
Task Introduction
you will be given a query, and two responses 
from two assistants, 
could you compare the two responses, 
and do the following: 
(1) write a concise review for each 
assistant's response, on how well the 
response answers the query, and whether 
it will be helpful to humans users, and any 
issues in the response;
(2) assigns a quantitative score on a scale 
of 10 to each response, reflecting 
your assessment of the two responses
Query: 
[query]
Response 1 from assistant 1: 
[response1]
Response 2 from assistant 2: 
[response2]
\end{verbatim}

\section{Appendix for Experimental settings}
\label{sec:appendix_exp_settings}

Here, we provide more details for experimental settings. 

\noindent\textbf{Hyper-parameters for the baseline PEFT methods} \quad For the P-tuning method, the soft prompts' length is 64, and the soft prompts is first initialized with dimension 36, and then a learnable projection layer projects it to the same dimension with the LlaMA-2 backbone. For P-tuning V2, the number of prompt tokens at each layer is set to 64. For LPT and IDPG, the bottleneck dimension is set to 1024, and the number of soft tokens is set to 4.   

For the adapter-based methods, Houlsby-Adapter and AdapterDrop, the bottleneck dimension is set to 18, and the adapter modules are added on the self-attention and feed-forward module. For the Parallel-Adapter and Learned-Adapter, the bottleneck dimension is set to 36, and the adapter modules are connected to the whole block. 

For LoRA, the initial rank at each module is set to 4. For AdaLoRA, the initial rank at each module is set to 8, and half of the rank budget is pruned during fine-tuning. 

We adjust the sparsity for SSP so that the number of tunable parameters is comparable with IAPT and the other baselines. For BitFit, the bias vectors are initialized with dimension 8, and then a learnable projection layer projects it to the same dimension with the LlaMA-2 backbone. For (IA)$^{3}$, the activation adjusting vectors are added the Query, Key, and Up activations. The adjusting vectors are initialized with dimension 16, and then a learnable projection layer projects it to the same dimension with the LlaMA-2 backbone.

\noindent\textbf{Training settings for PEFT methods} \quad We use the HugginFace Transformers \cite{wolf-etal-2020-transformers}, PEFT \cite{peft}, or the original code repositories for implementing all the methods, and for training and making predictions. For fine-tuning LlaMA-2 7B model, the maximum sequence length is set to 2048. The maximum training epoch is set to 10. The batch size is set between 16 for task with less than 10k training set, and 128 otherwise. We use AdamW as the optimizer with a linear learning rate decay schedule and 6\% of the training steps for warm-up. The learning rate is set to 1e-4. For the bi-level optimizaiton of IAPT, the validation set is the same with the dev set. The hyper-parameters for calculating the gradients of the architectural parameters are the same with the normal training procedure, except that the learning rate is 1e-6. The other hyper-parameters are kept the same with \cite{wolf-etal-2020-transformers}. In every 200 steps, the model is evaluated on the dev set. Patience is set to 10, that is, if the model does not achieve a lower development set loss for 10 evaluation runs, the training stops early. The best checkpoint on the dev set is used to run predictions on the test set.

\section{Appendix: settings for efficiency analysis}
\label{sec:instruction_for_efficiency_analysis}

In the Table \ref{tab:results_efficiency_analysis} of the main contents, we conduct analysis on the IAPT and other PEFT methods' memory and speed during inference. 

The example instruction we used in this analysis is presented below.

\begin{verbatim}
Generate a blog post of 500 words or less 
that discusses the following news article: 

The Department of Child Protection (DCP) 
must pay compensation and medical expenses 
to a youth worker who developed pericarditis 
after getting a Covid booster under a 
workplace vaccination directive, the South 
Australian Employment Tribunal has ruled.

In a decision handed down on 15 January 
2024, the Tribunal determined that Daniel 
Shepherd’s employment was “a significant 
contributing cause” to his injury, which 
has since rendered him incapable of 
performing his role at work.

Shepherd got a Covid booster in February 
2022 as a requirement for his ongoing 
employment with the DCP. The DCP admitted 
that Shepherd’s pericarditis had been 
caused by the booster, but denied 
responsibility for the injury, arguing that 
it did not arise from Shepherd’s employment, 
but from a lawful State Government Public 
Health Order (PHO), issued under the 
Emergency Management Act 2004 (EMA).
\end{verbatim}

We restrict the number of newly generated tokens to be 32 under the method of beam search with beam size equal to 1 or 3. The length of the initial instruction is 278 after adding the soft prompts and special tokens under the IAPT method, and 274 under the LoRA method. The LLM backbone is LlaMA-2 7B model. We run the generation process for 100 times to calculate the average metric values, reducing the randomness.

\section{Appendix: pilot experiments}
\label{sec:appendix_pilot_experiments}

We now conduct pilot experiments on the BoolQ and E2E tasks to demonstrate the necessity of learning activation functions for the prompt generators. The other hyper-parameters or experimental settings are kept the same with Section \ref{subsec:experimental_settings} and \ref{sec:appendix_exp_settings}. 

We now compare three variants of IAPT: (a) IAPT-relu, which is to set the activation function of the prompt generators to ReLU. (b) IAPT-relu, which is to set the activation function of the prompt generators to GeLU. (c) IAPT-relu-gelu, which is to set the activation functions of the prompt generators on the lower 16 Transformer layers to ReLU, and set those on the 16 higher Transformer layers to GeLU. (d) IAPT-gelu-relu, which is to set the activation functions of the prompt generators on the lower 16 Transformer layers to GeLU, and set those on the 16 higher Transformer layers to ReLU. The results on the BoolQ and E2E tasks are presented in Table \ref{tab:appendix_pilot_exp}. 

The results demonstrate that: (a) different downstream tasks may favor different activation functions for the prompt generators. (b) applying different activation functions for different Transformer layers may result in performance gains. The results demonstrate that there is room for improvements if we set the prompt generators' activation functions properly. However, such a setting is intractable to be set manually.

\begin{table}[tb!]
\centering
\resizebox{0.46\textwidth}{!}{
\begin{tabular}{c|ccc}
\hline
\multirow{2}*{\textbf{Method}}    &     \textbf{BoolQ}     &   \textbf{E2E}   &    \textbf{SQuAD}  \\ 

&    \textbf{(acc)}  &   \textbf{(rouge-l)}   &   \textbf{(f1-em)}  \\
\hline
IAPT   &     \textbf{87.5}    &    \textbf{71.3}   &     \textbf{88.5}       \\
IAPT-gelu   &   86.7   &   70.6   &   87.8     \\
IAPT-relu   &    86.4  &  70.7  &  87.7       \\
IAPT-relu-gelu   &    86.8    &  70.7  &  88.0      \\
IAPT-gelu-relu   &    86.6   &  70.5   &  87.9       \\
\hline
\end{tabular}}

\caption{\label{tab:appendix_pilot_exp} The experimental results for the pilot experiments. The backbone model is LlaMA-2 7B. } 
\end{table}

\section{Ablation on the soft prompt length} 
\label{sec:appendix_ablation_prompt_length}

\begin{figure}
\centering
\includegraphics[width=0.48\textwidth]{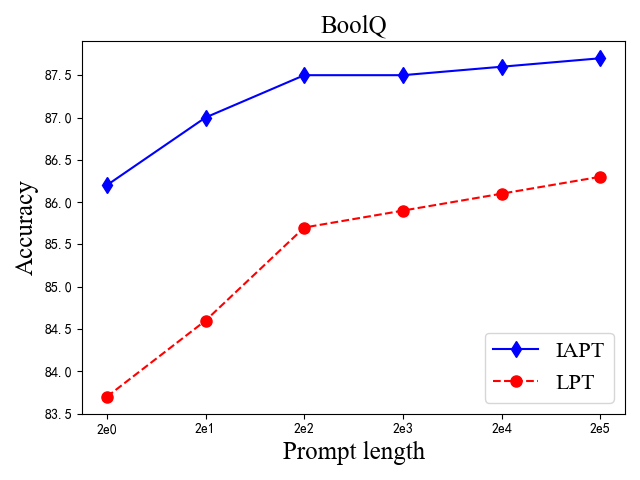}
\caption{Performances under different soft prompt lengths.  }
\label{fig:BoolQ_different_length}
\end{figure}

We vary the prompt length $l_{sp}$ from 4 to \{1, 2, 8, 16, 32\} for IAPT and LPT, and present the results on the BoolQ task in Figure \ref{fig:BoolQ_different_length}.

\section{Ablation on the pretrained backbones}

Our main experiments are conducted on the LlaMA-2 7B model. To demonstrate that our method works well regardless of the backbone models, we now conduct experiments on the GPT-2 large (774M parameters) and Pythia-1.4b models. The other experimental settings are kept the same with the main experiments (Table \ref{tab:results_main_1}). We conduct experiments on the BoolQ, E2E and SQuAD tasks. The results are reported in Table \ref{tab:results_different_backbones}.

\begin{table}[tb!]
\centering
\resizebox{0.43\textwidth}{!}{
\begin{tabular}{c|ccc}
\hline
\multirow{2}*{\textbf{Method}}    &     \textbf{BoolQ}     &   \textbf{E2E}   &    \textbf{SQuAD}  \\ 

&    \textbf{(acc)}  &   \textbf{(rouge-l)}   &   \textbf{(f1-em)}  \\
\hline 

\multicolumn{4}{c}{\textbf{\emph{Results for GPT-2 large}}}  \\
\hline

LPT   &     67.2    &   66.1     & 43.2 \\
LoRA   &        67.5    &   66.6     &  43.8    \\
AdaLoRA   &     67.6    &   66.5     & 43.7    \\
\hdashline
IAPT   &   \textbf{68.3}   &  \textbf{67.2}  &     \textbf{44.6}     \\

\hline 
\multicolumn{4}{c}{\textbf{\emph{Results for Pythia-1.4b large}}}  \\
\hline

LPT   &       71.3    &     68.2    &   50.1   \\
LoRA   &        71.6    &    68.4    &    50.4   \\
AdaLoRA   &     71.5    &   68.5     &   50.4  \\ 
\hdashline
IAPT   &    \textbf{72.3}   &  \textbf{69.1}  &     \textbf{51.3}     \\

\hline
\end{tabular}}
\caption{\label{tab:results_different_backbones} Results for different PEFT methods on the BoolQ, E2E and SQuAD benchmarks. The backbone LMs are GPT2-large and Pythia-1.4b.}
\end{table}

\section{Visualization of the learned activation functions}
\label{sec:appendix_visualize_activation}

\begin{figure*}[ht]	
\centering
\subfigure[1st to 4-th layer]{%
\includegraphics[width=0.42\textwidth]{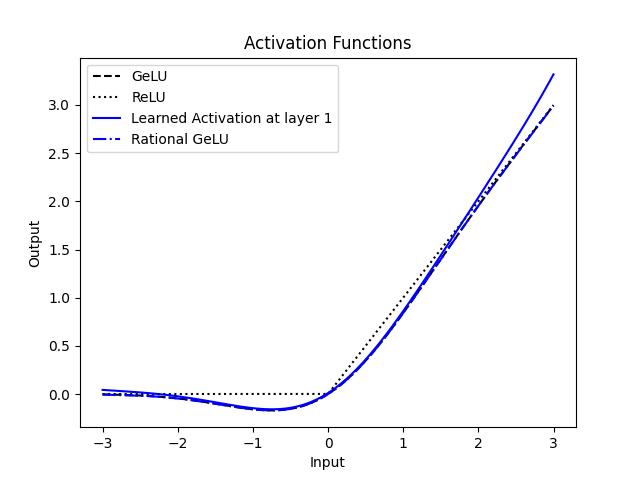}
\label{subfig:activation_funct_layer_1}
}
\subfigure[5th to 8-th layer]{%
\includegraphics[width=0.42\textwidth]{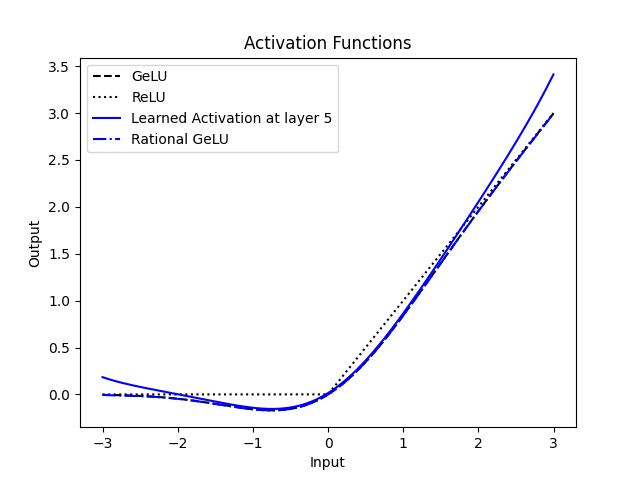}
\label{subfig:activation_funct_layer_5}
}
\subfigure[9th to 12-th layer]{%
\includegraphics[width=0.42\textwidth]{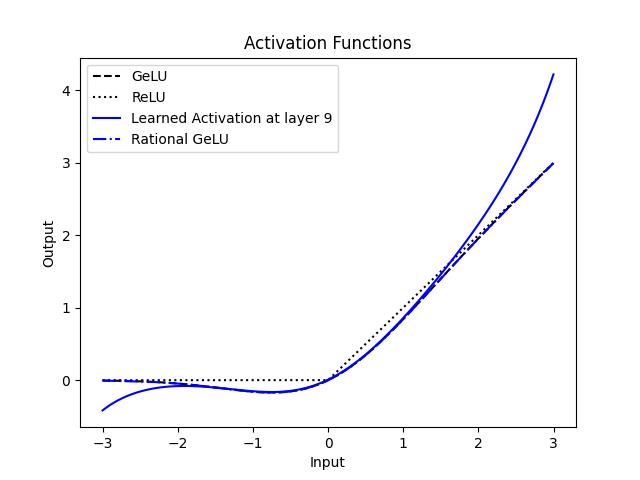}
\label{subfig:activation_funct_layer_9}
}
\subfigure[13th to 16-th layer]{%
\includegraphics[width=0.42\textwidth]{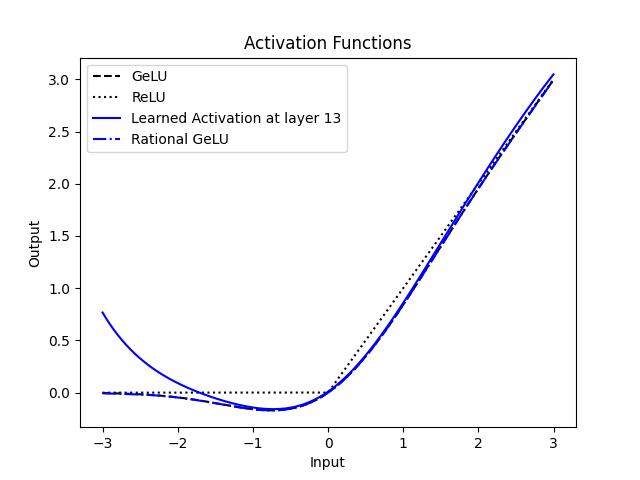}
\label{subfig:activation_funct_layer_13}
}
\subfigure[17th to 20-th layer]{%
\includegraphics[width=0.42\textwidth]{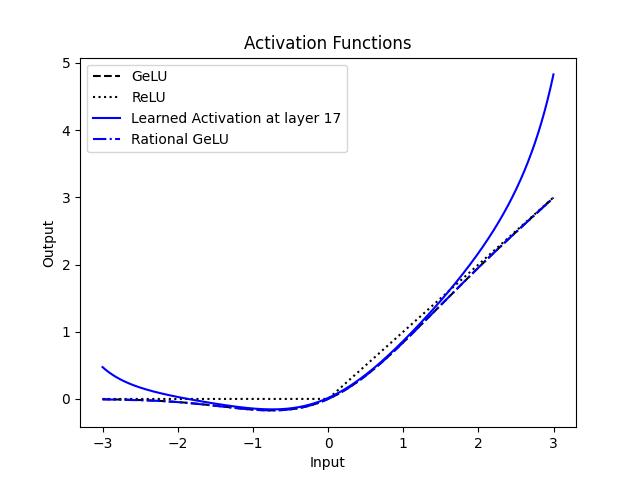}
\label{subfig:activation_funct_layer_17}
}
\subfigure[21th to 24-th layer]{%
\includegraphics[width=0.42\textwidth]{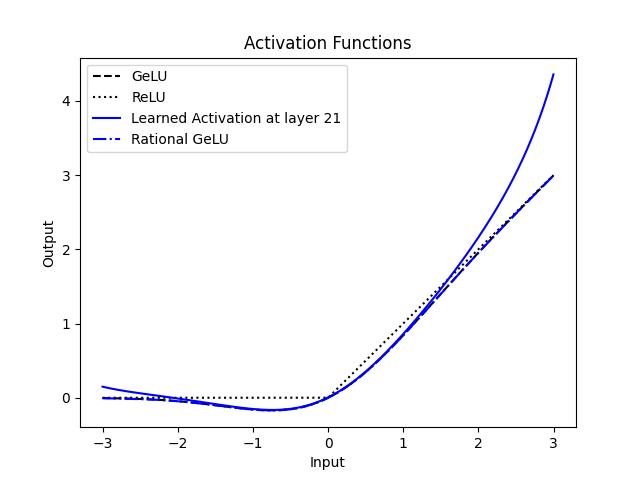}
\label{subfig:activation_funct_layer_21}
}
\subfigure[25th to 28-th layer]{%
\includegraphics[width=0.42\textwidth]{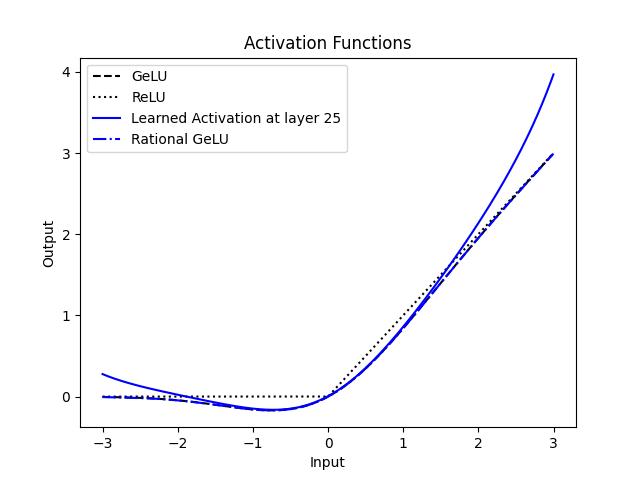}
\label{subfig:activation_funct_layer_25}
}
\subfigure[29th to 32-th layer]{%
\includegraphics[width=0.42\textwidth]{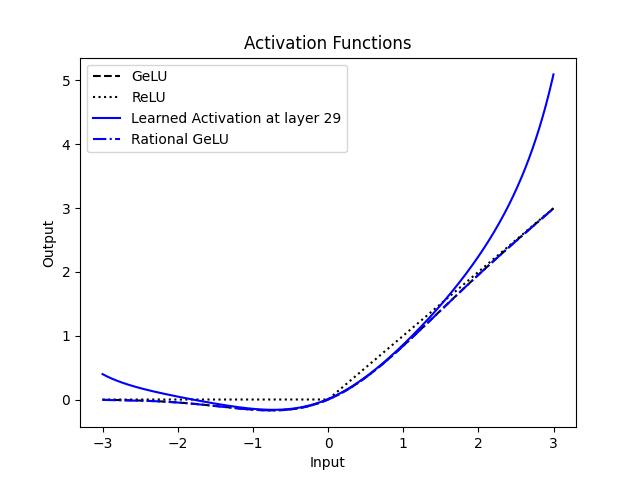}
\label{subfig:activation_funct_layer_29}
}
\caption{The learned activation functions for the prompt generators at different Transformer layers. }
\label{fig:activation_funct}

\end{figure*}

Now, we visualize the learned activation functions of the prompt generator at different Transformer layers in Figure \ref{fig:activation_funct}.

\section{Case studies of Instruction tuning}
\label{sec:appendix_case_study}

In the Section \ref{subsec:main_results} of the main content, we present the overall performance of IAPT and AdaLoRA on the MT-Bench, after fine-tuning LlaMA-2 7B on the Alpaca dataset. Now we present concrete examples in Table \ref{tab:results_alpaca_examples_1} to showcase the Superiority of IAPT.

\begin{table*}[ht]
\centering
\resizebox{0.98\textwidth}{!}{
\begin{tabular}{p{2.0cm}|p{12cm}}
\hline

Query &   \makecell[l]{One morning after sunrise, Suresh was standing facing a pole. The shadow of \\the pole fell exactly to his right. Can you tell me the direction towards which \\the shadow was pointing - east, south, west, or north? Explain your reasoning \\steps.} \\
\hline

\multicolumn{2}{c}{\emph{IAPT's response}}  \\
\hline
Responses  &   \makecell[l]{The shadow of the pole fell exactly to Suresh's right, which means that the \\shadow was pointing towards his left. Since the sun rises in the east and sets in \\the west, the shadow of the pole was pointing towards the west. Therefore, the \\direction towards which the shadow was pointing was west.}  \\  
\hdashline
GPT-4's comments     &  \makecell[l]{Assistant 1 provides a clear and logical explanation. They correctly identify \\that since the shadow falls to the right of Suresh, it means that he is facing \\north, because the sun rises in the east and its light comes from that direction, \\casting shadows towards the west. The assistant's reasoning is accurate \\considering that sunrise has been mentioned, thus we can deduce that Suresh \\must be facing north for the shadow to fall on his right side (west). The \\explanation is easy to understand and would be helpful to human users looking \\for a clear answer with justification. However, there is an internal contradiction \\in Assistant 1's response; initially they state "the shadow was pointing towards \\his left" which does not align with their correct conclusion or premise.\\\\Score: 7/10 - While Assistant 1 reaches the correct conclusion (that the shadow \\points west), they mistakenly describe how shadows fall relative to Suresh’s \\position first (claiming it falls towards his left), which could confuse users.}   \\
\hdashline
GPT-4 score   &    7    \\
\hline

\multicolumn{2}{c}{\emph{AdaLoRA's response}}  \\
\hline
Responses  &   \makecell[l]{The shadow of the pole fell exactly to Suresh's right, indicating that the sun was \\rising in the east.}  \\  
\hdashline
GPT-4's comments     &  \makecell[l]{Assistant 2 starts well by indicating where Suresh must be looking based on \\where his shadow falls but fails to complete their reasoning. They point out \\correctly that if a pole's shadow falls exactly to someone’s right after sunrise, \\it indicates that person must be facing north with sunlight coming from their \\left (east), casting a shadow toward their right (west). However, Assistant 2 \\does not provide this final step of information—identifying explicitly what \\direction this would make the shadow point—which makes their response \\incomplete when answering directly what was asked in terms of cardinal \\directions.\\\\Score: 5/10 - Assistant 2 identifies an important piece of information regarding \\how shadows are cast at sunrise but doesn't fully answer what direction toward \\which the shadow points nor do they offer any steps of reasoning beyond noting \\sun placement at sunrise. This may leave human users somewhat unclear about \\how they've reached any implicit conclusions or if additional reasoning should \\follow.}   \\
\hdashline
GPT-4 score   &    5   \\
\hline

\end{tabular}}
\caption{\label{tab:results_alpaca_examples_1} An example showcasing that the fine-tuning quality of IAPT is higher than AdaLoRA.  }
\end{table*}

\end{CJK*}

\end{document}